\PassOptionsToPackage{portrait}{geometry}
\documentclass[]{bytedance_seed}

\usepackage[toc,page,header]{appendix}

\usepackage{amsmath}
\usepackage{amssymb}
\usepackage{tikz}
\usetikzlibrary{shapes,arrows,positioning,fit,backgrounds,calc,shadows,decorations.pathreplacing}

\title{Generative molecule evolution using 3D pharmacophore for efficient Structure-Based Drug Design}

\author[1,\dagger, \circ]{Yi He}
\author[1, \circ]{Ailun Wang}
\author[1]{Zhi Wang}
\author[1]{Yu Liu}

\author[1]{Xingyuan Xu}
\author[1, \dagger]{Wen~Yan}

\affiliation[1]{ByteDance Seed}

\contribution[\dagger]{Corresponding authors}
\contribution[\circ]{Equal Contribution.}

\abstract{
Recent breakthroughs in generative models have shown great promise for drug discovery, yet Structure-Based Drug Design (SBDD) faces two critical challenges: the scarcity of high-quality protein-ligand complex data and the practical synthesizability of generated molecules. While ligand-based approaches benefit from billion-scale small molecule datasets, they often lack the structural context of the target protein. To address these dual challenges of data scarcity and synthetic inaccessibility, we propose \textbf{MEVO}, a generative molecular evolutionary framework that operates within a synthesizable chemical space. MEVO introduces a reaction-aware generative model trained on reaction routes constructed by applying predefined reaction templates to Enamine Building Blocks. The model jointly generates molecules and corresponding synthesis sketches consistent with these templates. By integrating this generative capability into a pocket-aware evolutionary strategy, MEVO iteratively refines molecular structures within the synthesizable chemical space, guided by a hierarchical scoring function that combines fast approximation with rigorous Absolute Binding Free Energy Perturbation (ABFEP) calculations. This framework efficiently generates high-affinity, synthetically feasible binders for diverse protein targets. We validate MEVO's performance through rigorous ABFEP calculations, demonstrating its ability to design potent inhibitors for challenging targets such as KRAS$^{\textrm{G12D}}$, achieving affinities comparable to known active inhibitors. By integrating the data richness of ligand-based learning with the precision of structure-based optimization, MEVO offers a holistic solution that ensures both high binding affinity and synthetic accessibility, paving the way for more efficient and practical structure-based drug design.
}

\date{\today}
\correspondence{Yi He at \email{heyi@bytedance.com}, Wen Yan at \email{wen.yan@bytedance.com}}

\begin{document}
\maketitle


\section{Introduction}

Drug discovery is a complex and iterative process, aiming to design molecules with sufficient affinity for specific protein targets while fulfilling stringent property requirements, such as pharmacokinetics, toxicity profile, and synthesizability. Among these, synthesizability serves as the fundamental prerequisite for experimental validation, while binding affinity remains the primary indicator of efficacy. In this work, we specifically address these two critical aspects to accelerate the design-make-test cycle. Traditional methodologies typically initiate with high-throughput screening (HTS) to identify starting compounds from predefined libraries~\cite{macarron_impact_2011, blay_high_2020}. To accelerate this process, virtual screening (VS) has become a standard alternative, employing empirical scoring functions~\cite{trott2010autodock, friesner2004glide, wang2002xscore} or machine learning models~\cite{stepniewska2018development, ballester2010rfscore, ozturk2018deepdta, jimenez2018kdeep, mcnutt2021gnina, wohlwend2025boltz2} to scan extensive databases. However, VS entails a critical speed-accuracy trade-off: approximate scoring functions, while fast, are prone to high false positive rates. Conversely, while Free Energy Perturbation (FEP) methods~\cite{wang2015accurate, cournia2017free} offer gold-standard accuracy, their high computational cost precludes their use in broad screening campaigns. Furthermore, even efficient screening methods are fundamentally bound by the limited chemical diversity of existing libraries. The theoretical drug-like space ($10^{33}$--$10^{60}$)~\cite{bohacek1996art, polishchuk2013estimation} far exceeds any enumerated collection, and even with the advent of trillion-scale synthesizable libraries~\cite{neumann_explore_2023}, efficiently navigating this vast landscape remains a formidable challenge.

The revolution of deep generative models in fields like computer vision and natural language processing has been supported by massive-scale datasets~\cite{kaplan2020scaling, schuhmann2022laion, gao2020pile}. This data abundance, coupled with scalable architectures~\cite{vaswani2017attention, brown2020language, ho2020denoising}, has enabled the learning of complex distributions and the generation of high-fidelity content. Following this paradigm, structure-based drug design (SBDD) has seen a surge in generative approaches~\cite{schneuing2022structure, guan2023decompdiff, peng2022pocket2mol}, showing promise in designing molecules complementary to protein pockets. However, training on experimentally determined protein--ligand complexes is limited by the availability of paired structural data. Databases such as PDBbind~\cite{liu2014pdb} provide valuable complex structures and binding measurements, but these paired data are much less abundant than ligand-only molecular data.

Ligand-based drug design (LBDD) can draw on large compound libraries such as ZINC~\cite{sterling2015zinc} and Enamine REAL~\cite{enamine_real} without requiring a corresponding protein--ligand complex for each molecule. Classical ligand-based generative models~\cite{gomez2018automatic, olivecrona2017molecular, jin2018junction} demonstrate how molecular structure data alone can be used to learn chemical representations and generate new molecules. By removing the requirement for paired complex structures, ligand-only learning provides access to a substantially larger pool of potential training molecules. Although docking-derived data such as CrossDocked~\cite{francoeur_three-dimensional_2020} and the generated conformations used by HelixDock~\cite{jiang2023helixdock} can expand the data available for protein--ligand modeling, they do not replace experimentally determined complexes, and their quality depends on the accuracy of docking and scoring~\cite{su_comparative_2019}. This motivates learning molecular generation from abundant ligand-only data and introducing target-specific structural guidance through an evolutionary framework.

Beyond data scarcity, a critical challenge lies in connecting computational design with experimental realization. SBDD models use different generation mechanisms: Pocket2Mol~\cite{peng2022pocket2mol} grows molecules autoregressively, whereas TargetDiff~\cite{guan2023decompdiff} jointly generates continuous atomic coordinates and categorical atom types through an equivariant diffusion process. Both model the target pocket without explicitly constraining generation to chemical reaction pathways, so chemically valid outputs may still be difficult to synthesize. Retrosynthesis-based validation adopts a "generate-then-check" approach, screening freely generated molecules with planning tools~\cite{coley2019robot, lin2020automatic}. This can be inefficient when candidates exhibit high synthetic complexity or require convoluted reaction pathways~\cite{gao2020synthesizability}. Reaction-based generation incorporates chemical transformations into the search process~\cite{horwood2020molecular, korovina2020chembo}, restricting exploration to products reachable under the chosen reaction model. Methods such as SynFlowNet~\cite{cretu2024synflownet} and SynLlama~\cite{sun2025synllama} generate molecules together with synthesis pathways, but do not directly generate the final molecules' 3D coordinates. Their reaction-based representations therefore do not directly impose 3D geometric constraints during generation, motivating methods that combine synthesis-aware generation with explicit structural guidance.

To bridge these gaps, we propose \textbf{MEVO}, a generative molecular evolutionary framework that operates within a synthesizable chemical space. Fundamentally, MEVO reformulates the SBDD challenge: instead of struggling to learn a direct mapping from protein structure to ligand structure from sparse data, we treat SBDD as an optimization problem within a robust, synthesizable chemical space learned from abundant ligand data. Specifically, MEVO decouples the design process into two synergistic components:
\begin{itemize}
    \item \textbf{Synthesis-Aware Generative Model:} Defines a synthesizable chemical space using small reactants and reliable reaction templates. This design constrains generation to reaction-reachable structures under template logic and enables the model to leverage abundant ligand data to learn robust chemical rules.
    \item \textbf{Structure-Guided Evolution:} Efficiently navigates the synthesizable chemical space to identify high-affinity binders. This decoupling allows for flexible guidance, where we introduce a hierarchical strategy utilizing fast surrogate scores for exploration and rigorous free energy calculations for precision.
\end{itemize}

We validate MEVO on diverse protein targets, demonstrating its capability to generate high-affinity binders while constraining candidates to template-consistent disconnections and short synthesis sketches. Our results highlight the effectiveness of MEVO in identifying potent, synthetically accessible candidates, offering a practical solution for data-efficient structure-based drug design.


\section{Results}
\label{sec:results}

\subsection{Unbiased Generation: Exploring the Drug-like Chemical Space}
\label{sec:results_generative_quality}
To assess the intrinsic generative capability of MEVO, we analyzed the physicochemical properties and diversity of 1,160 unfiltered valid molecules generated using shape-only conditioning. For this evaluation, we generated an independent set by explicitly targeting 200 molecules per target across the six targets (100 from 2-step reactions and 100 from 3-step reactions). Only the final products were recorded to evaluate the model's pure exploratory capability without the directional bias imposed by iterative evolutionary selection. The final yield of 1,160 molecules out of the targeted 1,200 reflects a minimal generative failure rate of $\sim$3.3\%, demonstrating the robustness of the reaction-based generation process.

\begin{figure}[htbp]
    \centering
    \includegraphics[width=0.95\textwidth]{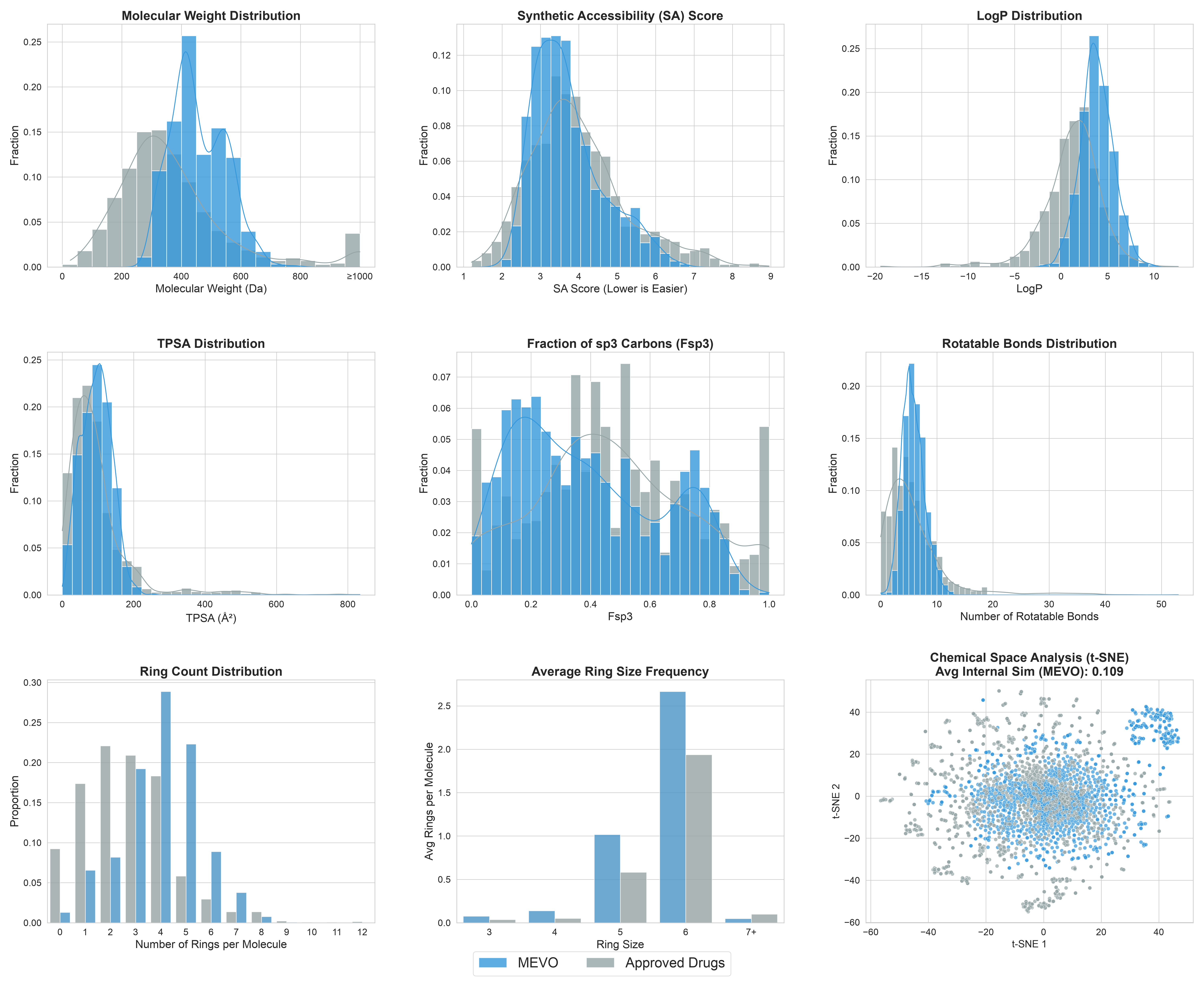}
    \caption{Intrinsic generative quality and chemical space analysis of MEVO-generated molecules compared to the ClinTox benchmark set. The 3$\times$3 grid presents: \textbf{Row 1} (Size, Synthesis \& Lipophilicity): Molecular Weight (MW), Synthetic Accessibility (SA) Score, and LogP; \textbf{Row 2} (Polarity, Complexity \& Flexibility): Topological Polar Surface Area (TPSA), Fraction of sp3 Carbons (Fsp3), and Rotatable Bonds; \textbf{Row 3} (Ring Systems \& Global Diversity): Ring Count, Ring Size distributions, and a t-SNE projection of Morgan fingerprints. The generated molecules (blue) exhibit properties largely overlapping with approved drugs (red) and broadly cover the biologically relevant chemical space, demonstrating high drug-likeness, structural complexity, and synthetic feasibility.}
    \label{fig:chemical_space}
\end{figure}

It is important to emphasize that the distributions shown in Figure~\ref{fig:chemical_space} represent the \textbf{raw generated molecules} without applying any filters (neither SMARTS-based structural alerts nor RO5 criteria). This distinction is crucial as it demonstrates the model's intrinsic capability to sample drug-like chemical space without relying on post-generation filtering to artificially improve distribution statistics. In contrast, for the binding affinity and synthetic accessibility analyses (discussed in Section~\ref{sec:sbdd_setup}), we apply chemical structural-alert filters and target-specific soft penalties to prioritize candidates for further validation.

Figure~\ref{fig:chemical_space} compares the property distributions and overall chemical space of these MEVO-generated molecules against a benchmark set of 1,386 FDA-approved drugs sourced from the ClinTox dataset~\citep{wu2018moleculenet}. This ClinTox benchmark set provides a baseline for drug-like physicochemical properties and structural motifs.

The results demonstrate that MEVO generates molecules with physicochemical profiles largely consistent with approved drugs. A detailed examination reveals a slight rightward shift in the Molecular Weight (MW) distribution (Row 1, left) compared to the ClinTox benchmark. This deviation is expected and design-driven: targets like KRAS$^{\textrm{G12D}}$ possess large, extended binding pockets, and the reference ligands used for shape conditioning are correspondingly large (MW $> 500$). Since MEVO respects the volumetric constraints of the target, the generated molecules naturally reflect these dimensions.

Despite the shift toward larger molecules, MEVO retains favorable synthetic complexity profiles. As shown in the SA Score distribution (Row 1, middle), MEVO molecules achieve scores comparable to or better than approved drugs. Beyond this heuristic metric, MEVO constrains candidates to be consistent with reaction templates and provides an explicit synthesis sketch (disconnections and reaction-set constraints) for every generated molecule. These template-consistent routes typically involve only 2--3 steps from small building blocks, and serve as a practical starting point for experimental refinement; detailed schemes for high-affinity candidates are provided in Appendix~\ref{sec:synthesis_details}.

We further analyzed the compliance with Lipinski's Rule of 5 (RO5)~\citep{lipinski1997experimental}. Among the generated molecules, 57.0\% fully comply with RO5 (0 violations), and 79.7\% have $\le 1$ violation. The violations are primarily driven by Molecular Weight ($>500$ Da, 33.8\% of molecules) and LogP ($>5$, 24.9\% of molecules), consistent with the target-driven size distribution discussed above. Given that many modern drugs, especially for difficult targets with shallow or cryptic pockets (e.g., KRAS$^{\textrm{G12D}}$) or those targeting protein-protein interactions (PPIs), operate in the "Beyond Rule of 5" (bRO5) space~\citep{doak2014how}, this profile is expected and acceptable, particularly when accompanied by high synthetic accessibility.

Furthermore, the distributions of Fsp3 and Rotatable Bonds (Row 2) indicate that MEVO avoids generating flat, aromatic-rich structures, instead producing candidates with appropriate 3D complexity and flexibility. The ring analysis (Row 3) further confirms this structural diversity: MEVO molecules typically contain 3--5 rings, with a balanced distribution of ring sizes, predominantly 5- and 6-membered rings, mirroring the structural patterns found in approved drugs. This suggests that the model successfully learns to assemble pharmacologically relevant scaffolds rather than simply concatenating acyclic fragments.

A key goal of \textit{de novo} design is to explore novel chemical space while remaining within the boundaries of drug-like properties. The t-SNE visualization (Row 3, right) demonstrates that MEVO molecules are not clustered in a single narrow region but are widely distributed across the chemical space, showing significant overlap with the space occupied by approved drugs. Quantitatively, the internal diversity of the generated set is high, with an average pairwise Tanimoto similarity of 0.109. This low similarity score indicates that the generated molecules are structurally distinct from one another, covering a broad range of chemotypes. This high diversity is particularly valuable in the early stages of drug discovery, as it provides a rich pool of candidates for screening against difficult targets. By effectively sampling the drug-like chemical space, MEVO increases the probability of finding novel hits that can be further optimized into lead compounds.

\subsection{Highly Potent and Synthesizable Candidates via Target-Specific Evolution}
\label{sec:sbdd_setup}

Having established the model's intrinsic capability to generate diverse and drug-like scaffolds, we deployed MEVO for Structure-Based Drug Design (SBDD). We evaluated the framework on six diverse protein targets spanning a wide range of binding site geometries and therapeutic classes: KRAS$^{\textrm{G12D}}$, TYK2, BACE, c-MET, HSP90, and p38. The set covers oncogenic GTPases (KRAS$^{\textrm{G12D}}$), kinases (TYK2, c-MET, p38), an aspartyl protease (BACE), and a chaperone ATPase (HSP90), exposing MEVO to cryptic pockets, shallow allosteric sites, deep enzymatic clefts, and ATP-binding sites alike. Among these, KRAS$^{\textrm{G12D}}$ is particularly notable as a historically challenging target due to its shallow, polar binding pocket and high sequence conservation.

To guide candidates toward biologically relevant and synthetically tractable chemical space, we applied target-specific soft penalties during evolution. These penalties favored desired physicochemical properties (e.g., MW, LogP, TPSA) and formal-charge profiles based on known active reference ligands. Negative charges were penalized across all targets, and the absence of a salt bridge with ASP12 was penalized for KRAS$^{\textrm{G12D}}$. All target-specific constraints were soft penalties rather than hard rejection criteria. A comprehensive breakdown of these preferences and the evolutionary hyperparameters (e.g., population size, parallel runs) is detailed in Appendix~\ref{sec:evolutionary_settings_main}.

Through parallel evolutionary trajectories, we identified the top 20 structurally distinct candidates for each target. To rigorously evaluate their binding affinity, we employed Absolute Binding Free Energy Perturbation (ABFEP), a highly accurate physics-based method. We first validated this protocol on known reference ligands, confirming its robust ability to correctly rank the compounds (Spearman $\rho$ generally exceeding $0.7$; full validation metrics are provided in Appendix~\ref{sec:abfep_validation}).

Before ABFEP evaluation, generated candidates were filtered using the SMARTS-based structural-alert rules (e.g., reactive electrophiles, PAINS~\citep{baell2010new}, unstable bonds; detailed in Appendix~\ref{sec:filtering_rules}). Target-specific physicochemical, charge, and interaction preferences remained soft penalties.

\begin{figure}[htbp]
    \centering
    \includegraphics[width=0.85\textwidth]{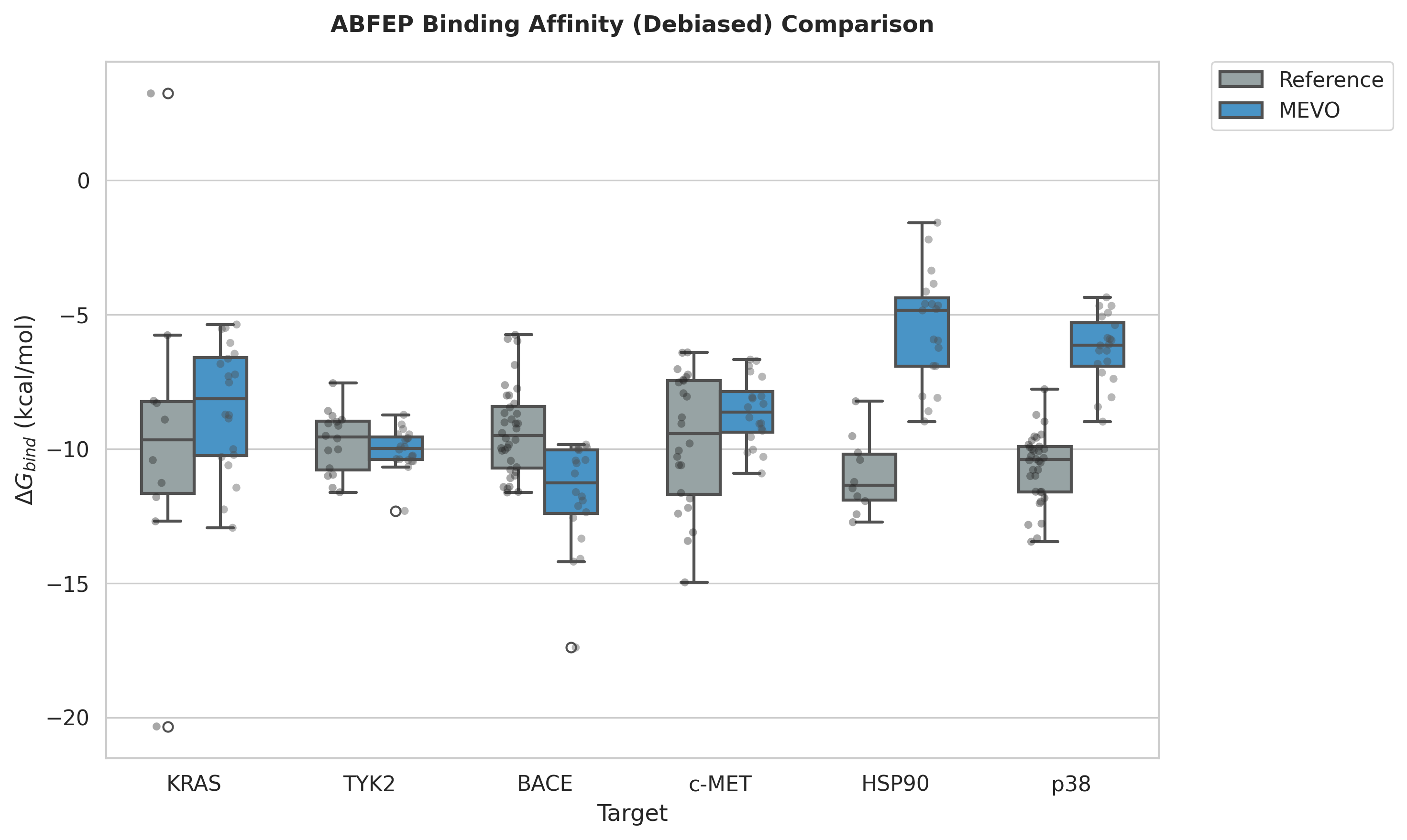}
    \caption{Comparison of debiased ABFEP binding affinities ($\Delta G_{bind}$) between reference ligands and the chemically filtered MEVO-generated candidates across multiple targets. Lower values indicate stronger binding affinity.}
    \label{fig:affinity_box}
\end{figure}

Figure~\ref{fig:affinity_box} presents the ABFEP affinity distributions, while Table~\ref{tab:affinity_summary} summarizes the best predicted candidate for each target. For each target, the reported ABFEP binding free energies of the reference ligands and MEVO-generated candidates were corrected using the same bias estimated from that target's reference ligands.

\begin{table}[htbp]
    \centering
    \caption{Summary of the best MEVO-generated candidates compared to reference ligands. The predicted $K_d$ is estimated from the debiased $\Delta G_{\text{bind}}$ at 298.15 K.}
    \label{tab:affinity_summary}
    \begin{tabular}{lcccc}
        \hline
        \textbf{Target} & \textbf{Best MEVO $\Delta G_{\text{bind}}$} & \textbf{Estimated MEVO $K_d$} & \textbf{Ref $\Delta G_{\text{bind}}$ Range} \\ \hline
        KRAS$^{\textrm{G12D}}$ & $-12.94$ kcal/mol & 327.7 pM & $[-20.34, 3.24]$ kcal/mol \\
        TYK2 & $-12.31$ kcal/mol & 948.2 pM & $[-11.61, -7.55]$ kcal/mol \\
        BACE & $-17.40$ kcal/mol & 0.2 pM & $[-11.62, -5.75]$ kcal/mol \\
        c-MET & $-10.91$ kcal/mol & 10.0 nM & $[-14.97, -6.40]$ kcal/mol \\
        HSP90 & $-8.98$ kcal/mol & 262.7 nM & $[-12.73, -8.23]$ kcal/mol \\
        p38 & $-8.98$ kcal/mol & 259.6 nM & $[-13.46, -7.78]$ kcal/mol \\
        \hline
    \end{tabular}
\end{table}

Examining the best candidates for each target (Table~\ref{tab:affinity_summary}), MEVO obtains sub-picomolar estimated affinity for BACE ($0.2$ pM) and sub-nanomolar estimated affinity for TYK2 ($948.2$ pM). The best BACE and TYK2 candidates have more favorable predicted binding free energies than the strongest reference ligands in their respective datasets. For KRAS$^{\textrm{G12D}}$, the top candidate also reaches sub-nanomolar estimated affinity at $327.7$ pM ($\Delta G_{\text{bind}} = -12.94$ kcal/mol).

While the best generated candidates for c-MET, HSP90, and p38 do not surpass the most potent reference ligands, their estimated $K_d$ values are $10.0$, $262.7$, and $259.6$ nM, respectively. Thus, each target has at least one MEVO-generated candidate with sub-micromolar predicted affinity. The reference datasets comprise highly optimized clinical candidates and approved drugs, representing the culmination of multi-year, resource-intensive lead optimization (LO) campaigns. Evaluating a \textit{de novo} hit discovery engine against these references is an exceptionally stringent benchmark. The best MEVO candidates provide promising starting points for subsequent SAR exploration.

\begin{figure}[htbp]
    \centering
    \includegraphics[width=0.85\textwidth]{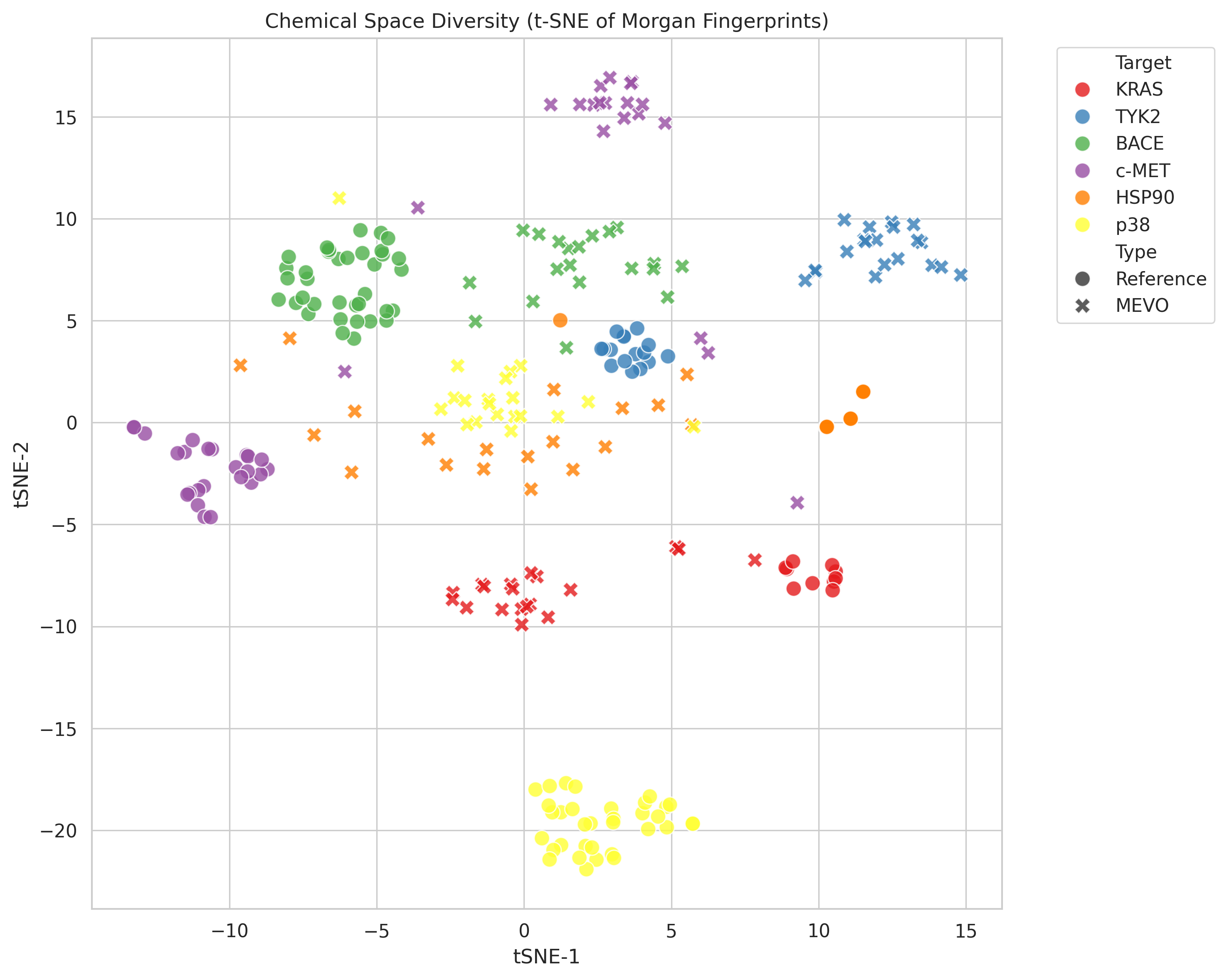}
    \caption{t-SNE projection of Morgan fingerprints comparing the chemical space of the top 20 MEVO-generated candidates against known reference ligands across all six targets. The generated molecules demonstrate significant structural novelty and internal diversity, exploring broader chemotypic spaces rather than clustering tightly around known scaffolds.}
    \label{fig:tsne_all_targets}
\end{figure}

Furthermore, to emphasize the generative novelty and internal diversity of MEVO's candidates, we visualized the chemical space distribution using a t-SNE projection of Morgan fingerprints (Figure~\ref{fig:tsne_all_targets}). The visualization reveals that MEVO-generated candidates (crosses) occupy distinct regions of the chemical space compared to the known reference ligands (circles). This separation demonstrates that the generative process successfully explores novel chemotypes and avoids simple analoging or memorization of known binders. More importantly, the internal diversity among the MEVO candidates themselves is exceptionally high. As measured by the average pairwise Tanimoto similarity, the MEVO-generated sets exhibit low internal similarity scores (ranging from 0.145 for HSP90 to 0.302 for KRAS$^{\textrm{G12D}}$). This indicates that rather than converging onto a single dominant scaffold, MEVO effectively explores a broad and diverse structural space, providing a rich pool of distinct starting points that is critical for maximizing success rates in early-stage hit discovery.

Crucially, this extreme potency and diversity were not achieved at the expense of practical viability. By operating intrinsically on a graph of valid chemical reactions, MEVO provides an explicit synthesis sketch---detailing the required reactants and reaction templates---for every generated candidate (examples of top-tier candidates are provided in Appendix~\ref{sec:synthesis_details}). However, we acknowledge a practical gap between these computational sketches and real-world bench experiments. First, the SMARTS-based reaction templates represent generalized reaction patterns rather than strict single-step experimental procedures; they may implicitly merge multiple steps or omit common reagents to maximize expressiveness for model learning. Second, evolution uses small reactants that may be more readily available, but supplier availability has not been verified. Finally, the presence of multiple reactive sites on a reactant may necessitate further experimental adjustments, such as the use of protecting groups. We do not explicitly resolve these experimental intricacies within the model; rather, we provide the computational routes as foundational sketches for synthetic chemists to refine. Importantly, these practical considerations typically influence only the specific experimental execution, not the fundamental synthesizability of the final molecule. Therefore, utilizing these reaction-based routes as constraints remains highly effective in ensuring the practical viability of the generated candidates.

\section{Discussion}
\label{sec:discussion}

The results presented above demonstrate that MEVO can effectively navigate the vast chemical space to identify high-affinity binders for challenging targets. A critical advantage of our framework lies in its synthesis-aware generation process, which constrains candidates to be consistent with reaction templates and provides template-consistent disconnections as synthesis sketches.

\subsection{Core Strengths}

\subsubsection{Synthesizability and Practical Application}
MEVO learns reaction-based generation from routes constructed using Enamine Building Blocks. Reaction-template constraints provide explicit synthesis sketches, in contrast to atom-based or topology-based generation that does not explicitly model synthesis routes. The use of small reactants during evolution may favor their availability, but supplier availability has not been verified. This reaction-based representation can be combined with different evolutionary scoring objectives.

Furthermore, our framework tends to generate concise synthesis sketches, typically involving only 1--3 reaction steps. This feature is practically advantageous: short routes reduce cumulative yield loss and simplify purification, lowering the experimental overhead required to realize candidates. Importantly, reaction templates encode disconnections and substructure-level feasibility, but do not model chemoselectivity, protecting group strategies, or detailed reaction conditions. In practice, experimental chemists can often resolve these gaps by applying standard edits such as protecting groups, order swapping, reagent equivalence, and condition optimization while preserving the core disconnections.

\subsubsection{Versatility in Design Objectives}
The integration of evolutionary strategies with deep generative models creates a powerful feedback loop. A key strength of MEVO is its flexible initialization strategy, which allows users to tailor the conditioning constraints to distinct drug discovery stages.

\begin{itemize}
    \item \textbf{Broad Exploration (Shape-only)}: By conditioning solely on the shape point cloud derived from a reference ligand (without specifying any atoms), the model performs unbiased exploration. This mode is particularly valuable for discovering novel binding modes or chemotypes that fit the pocket volume but interact through mechanisms distinct from the reference. While our current validation relies on reference-derived shapes, this approach is theoretically extensible to orphan targets (where no ligand is known) by defining the shape constraint using pocket detection algorithms or expert-guided volumetric definitions.
    \item \textbf{Scaffold Hopping}: When key pharmacophoric anchors (e.g., hydrogen bonding motifs) are enforced while the central core is left unconstrained, the model can traverse untested chemical space to find novel cores that maintain critical interactions. This enables the discovery of structurally distinct chemotypes while retaining biological activity.
    \item \textbf{R-group Optimization}: Conversely, by fixing a privileged scaffold and allowing evolution only on peripheral substituents, the framework can rapidly optimize binding affinity and physicochemical properties around a validated core.
\end{itemize}

This versatility is crucial for difficult targets like KRAS, where the design objective may shift from finding any binder (Broad Exploration) to optimizing a specific lead series (R-group Optimization).

\subsubsection{Extensibility}
The evolutionary framework relies on a scoring function to guide the generative process. MEVO employs a hierarchical strategy: computationally inexpensive surrogate scores (combining ByteFF energy and ProLIF interaction fingerprints) drive the rapid exploration of the inner loop, while rigorous but costly ABFEP calculations are reserved for the outer loop verification. This balance ensures that the model can evaluate thousands of candidates without prohibitive computational costs.

Crucially, the scoring architecture is designed to be modular and "plug-and-play." While this work focuses primarily on binding affinity and synthetic accessibility, the framework can be readily extended to incorporate additional objectives. For instance, machine learning models predicting ADMET properties (e.g., solubility, metabolic stability, toxicity) or selectivity profiles against off-targets can be inserted as additional terms in the surrogate scoring function. This extensibility allows MEVO to evolve from a pure affinity optimizer into a multi-objective drug design platform, capable of addressing the complex parameter space required for clinical success.

\subsection{Limitations and Practical Considerations}
Nevertheless, bridging the gap between computational design and experimental realization requires a nuanced understanding of the limitations inherent in template-based approaches. It is important to acknowledge that reaction templates operate primarily on substructure matching and do not explicitly model complex steric or electronic effects. For instance, a template for amide coupling will validly connect an amine and a carboxylic acid, but may not account for severe steric hindrance introduced by bulky ortho-substituents. Similarly, while we filter for general structural alerts, our current framework does not automatically design protecting group strategies. Therefore, we view the generated synthesis steps as editable sketches that constrain the final product to a reaction-reachable space rather than as step-by-step experimental SOPs. In practice, experienced chemists can often adapt these sketches by modifying conditions, employing alternative coupling reagents, or introducing protecting groups while preserving the key disconnections.

\subsection{Future Directions}
Future iterations of MEVO could incorporate data-driven reaction yield prediction models or retrosynthetic planning tools that explicitly account for protecting group strategies, further reducing the gap between \textit{in silico} design and \textit{in vitro} realization.

Additionally, while our current framework assumes a fixed receptor structure (conditioned on a specific shape), the evolutionary nature of MEVO opens the door to dynamic pocket modeling. By integrating a protein-ligand cofolding model, future iterations could co-evolve both the ligand and the receptor conformation, progressively uncovering cryptic binding sites or induced-fit states that are inaccessible to static SBDD methods.

\section{Conclusion}
\label{sec:conclusion}

In this work, we introduced \textbf{MEVO}, a synthesis-aware evolutionary generative framework designed to address the twin challenges of data scarcity and synthetic accessibility in structure-based drug design. By leveraging a discrete diffusion model trained on large-scale ligand datasets, MEVO internalizes fundamental chemical rules and reaction logic, pairing each generated molecule with a template-consistent synthesis sketch that provides disconnection constraints for experimental realization.

The integration of this generative capability into a pocket-aware evolutionary strategy allows for the systematic exploration of the synthesizable chemical space. Guided by a hierarchical scoring function that balances the speed of surrogate potential energy calculations with the accuracy of Absolute Binding Free Energy Perturbation (ABFEP), MEVO effectively bridges the gap between ligand-based data richness and structure-based precision. Our validation on challenging targets, particularly the successful design of potent inhibitors for KRAS$^{\textrm{G12D}}$, demonstrates the practical utility and robustness of this approach.

MEVO represents a significant step towards fully automated, synthetically grounded \textit{de novo} drug design. Future developments will focus on extending the framework to incorporate multi-objective optimization for ADMET properties and exploring its applicability to a broader range of complex biological targets. By unifying chemical feasibility with high-affinity optimization, MEVO provides a powerful and practical toolkit for accelerating the discovery of next-generation therapeutics.

\section{Methods}
\label{sec:methods}

\subsection{Data}
\label{subsec:data}

\subsubsection{Data Preparation}
We construct reaction trees containing one, two, or three reaction nodes. The
starting pool contains 152,296 unique, single-component compounds selected from
the Enamine Building Blocks collection. Each building block has at most 14 heavy
atoms and contains only B, C, N, O, F, Si, P, S, Cl, Br, or I. We apply 110
validated reaction SMARTS templates with RDKit~\cite{landrum2013rdkit}. Longer
routes are formed recursively by using a product from a shorter route in a
reactant slot of a new root reaction. Each accepted product is sanitized,
connected, and contains 15--50 heavy atoms, while the combined reactants of
each reaction contain at most 50 heavy atoms.

Reaction templates in our dataset operate at the level of substructure matching and encode disconnections rather than detailed experimental conditions. As a result, the resulting step sequences should be interpreted as template-consistent synthesis sketches that constrain the final product to a reaction-reachable space, while experimental realization may require standard synthesis edits such as protecting groups, order swapping, reagent equivalence, and condition optimization.

Each retained route is paired with one final-product conformer generated with
ETKDGv3 and minimized with MMFF94~\cite{halgren1996merck}. The stored
coordinates contain heavy atoms only and are centered at their geometric
centroid; reactants and intermediate products have no coordinates. Routes for
which conformer generation or force-field parameterization fails are
discarded. The resulting training data contain 155,379,402 one-step routes,
185,356,061 two-step routes, and 147,412,580 three-step routes.

\subsubsection{Representation}
We use a matrix representation of atomic properties, molecular connectivity,
pharmacophores, and final-product geometry.

Each chemical entity (a product molecule or a set of reactant molecules) is represented as a fixed-size matrix $M \in \mathbb{Z}^{50 \times 64}$. Multiple reactant molecules are concatenated along the row dimension, and all inputs are padded to 50 rows. All columns use a shared vocabulary of 102 token IDs. Token 0 represents a zero or padding value, token 101 is reserved for MASK corruption, and feature-specific masks restrict each column to its valid categories. The 64 feature dimensions are organized into four complementary blocks:

\begin{itemize}
    \item \textbf{Atomic attributes (Columns 1--3)}: Atomic number, number of attached hydrogens, and formal charge.
    \item \textbf{Pharmacophores (Columns 4--8)}: Five independent binary indicators for hydrogen-bond acceptor, aromatic, hydrogen-bond donor, negatively ionizable, and positively ionizable features.
    \item \textbf{Topology (Columns 9--58)}: Encodes molecular connectivity by embedding the adjacency matrix, with bond types represented as categorical values. These 50 columns correspond exactly to the maximum capacity of 50 heavy atoms, where each column represents the bond type to a specific atom index (0--49). For padded positions (where no atom exists) or pairs with no chemical bond, the entries are assigned a special padding token (zero). This explicit representation enables the model to learn complex structural transformations during synthesis.
    \item \textbf{3D geometry (Columns 59--64)}: Each Cartesian coordinate is represented by a coarse value $c\in\{0,\ldots,99\}$ and a fine value $f\in\{0,\ldots,9\}$, stored as token IDs $c+1$ and $f+1$, respectively. The coordinate is decoded as $c+0.1f-50$~\AA{}. This gives 1,000 values from $-50.0$ to $49.9$~\AA{} at 0.1~\AA{} resolution, with a maximum quantization error of 0.05~\AA{} per axis.
\end{itemize}

\begin{figure*}[ht]
\centering
\begin{tikzpicture}[
    font=\sffamily\small,
    scale=0.85, transform shape
]

\definecolor{inputblue}{HTML}{2E5AA8}
\definecolor{condgreen}{HTML}{2E8B57}
\definecolor{transpurple}{HTML}{7B68EE}
\definecolor{outputorange}{HTML}{D2691E}
\definecolor{bordergray}{HTML}{888888}

\colorlet{atomattr}{inputblue!10}
\colorlet{pharma}{condgreen!10}
\colorlet{topo}{transpurple!10}
\colorlet{geom}{outputorange!10}

\colorlet{accentblue}{inputblue}

\def\wAtom{1.6}
\def\wPharma{2.0}
\def\wTopo{6.0}
\def\wGeom{2.0}
\def\H{1.6}
\def\W{\wAtom + \wPharma + \wTopo + \wGeom}
\def\gap{0.5}

\begin{scope}[shift={(0, \H+\gap)}]
    \fill[white] (0,0) rectangle (\W, \H);
    
    \fill[atomattr] (0,0) rectangle (\wAtom, \H);
    \fill[pharma] (\wAtom,0) rectangle (\wAtom+\wPharma, \H);
    \fill[topo] (\wAtom+\wPharma,0) rectangle (\wAtom+\wPharma+\wTopo, \H);
    \fill[geom] (\wAtom+\wPharma+\wTopo,0) rectangle (\W, \H);
    
    \draw[accentblue, thick, rounded corners=2pt] (0,0) rectangle (\W, \H);
    \draw[bordergray] (\wAtom,0) -- (\wAtom,\H);
    \draw[bordergray] (\wAtom+\wPharma,0) -- (\wAtom+\wPharma,\H);
    \draw[bordergray] (\wAtom+\wPharma+\wTopo,0) -- (\wAtom+\wPharma+\wTopo,\H);
    
    \node[anchor=east, font=\tiny, text=gray] at (-0.1, \H-0.15) {Atom 1};
    \node[anchor=east, font=\tiny, text=gray] at (-0.1, \H/2) {$\vdots$};
    \node[anchor=east, font=\tiny, text=gray] at (-0.1, 0.15) {Atom 50};
    
    \node[anchor=east, font=\bfseries, text=accentblue] at (-1.2, \H/2) {Reactants};
    \node[anchor=east, font=\tiny, text=gray] at (-1.2, \H/2-0.22) {$50 \times 64$};
    
    \node[font=\scriptsize, text=inputblue, align=center] at (\wAtom/2, \H/2) {Atomic\\Attrs};
    \node[font=\scriptsize, text=condgreen, align=center] at (\wAtom+\wPharma/2, \H/2) {Pharmaco-\\phores};
    \node[font=\scriptsize, text=transpurple, align=center] at (\wAtom+\wPharma+\wTopo/2, \H/2) {Topology (Adjacency Matrix)};
    \node[font=\scriptsize, text=outputorange, align=center] at (\wAtom+\wPharma+\wTopo+\wGeom/2, \H/2) {3D\\Coords};
\end{scope}

\begin{scope}[shift={(0, 0)}]
    \fill[white] (0,0) rectangle (\W, \H);
    
    \fill[atomattr] (0,0) rectangle (\wAtom, \H);
    \fill[pharma] (\wAtom,0) rectangle (\wAtom+\wPharma, \H);
    \fill[topo] (\wAtom+\wPharma,0) rectangle (\wAtom+\wPharma+\wTopo, \H);
    \fill[geom] (\wAtom+\wPharma+\wTopo,0) rectangle (\W, \H);
    
    \draw[accentblue, thick, rounded corners=2pt] (0,0) rectangle (\W, \H);
    \draw[bordergray] (\wAtom,0) -- (\wAtom,\H);
    \draw[bordergray] (\wAtom+\wPharma,0) -- (\wAtom+\wPharma,\H);
    \draw[bordergray] (\wAtom+\wPharma+\wTopo,0) -- (\wAtom+\wPharma+\wTopo,\H);
    
    \node[anchor=east, font=\tiny, text=gray] at (-0.1, \H-0.15) {Atom 1};
    \node[anchor=east, font=\tiny, text=gray] at (-0.1, \H/2) {$\vdots$};
    \node[anchor=east, font=\tiny, text=gray] at (-0.1, 0.15) {Atom 50};
    
    \node[anchor=east, font=\bfseries, text=accentblue] at (-1.2, \H/2) {Product};
    \node[anchor=east, font=\tiny, text=gray] at (-1.2, \H/2-0.22) {$50 \times 64$};
\end{scope}


\draw[<->, thick, accentblue] (0, -0.5) -- (\W, -0.5);
\node[font=\small, fill=white, inner sep=2pt] at (\W/2, -0.5) {64 Features};

\node[font=\tiny, text=gray] at (\wAtom/2, -0.9) {Cols 1--3};
\node[font=\tiny, text=gray] at (\wAtom+\wPharma/2, -0.9) {Cols 4--8};
\node[font=\tiny, text=gray] at (\wAtom+\wPharma+\wTopo/2, -0.9) {Cols 9--58};
\node[font=\tiny, text=gray] at (\wAtom+\wPharma+\wTopo+\wGeom/2, -0.9) {Cols 59--64};

\draw[decorate, decoration={brace, amplitude=8pt, mirror}, thick, accentblue] 
    (\W+0.25, -0.05) -- (\W+0.25, 2*\H+\gap+0.05);
\node[align=left, font=\small, anchor=west] at (\W+0.5, \H+\gap/2) {
    \textbf{One Synthesis Step}\\
    \textcolor{gray}{\footnotesize $100 \times 64$}
};

\begin{scope}[shift={(0, -1.5)}]
    \fill[atomattr, draw=bordergray, thin] (0,0) rectangle (0.35, 0.22);
    \node[anchor=west, font=\scriptsize] at (0.45, 0.11) {Atomic Attributes};
    
    \fill[pharma, draw=bordergray, thin] (3.0,0) rectangle (3.35, 0.22);
    \node[anchor=west, font=\scriptsize] at (3.45, 0.11) {Pharmacophores};
    
    \fill[topo, draw=bordergray, thin] (6.2,0) rectangle (6.55, 0.22);
    \node[anchor=west, font=\scriptsize] at (6.65, 0.11) {Topology};
    
    \fill[geom, draw=bordergray, thin] (8.3,0) rectangle (8.65, 0.22);
    \node[anchor=west, font=\scriptsize] at (8.75, 0.11) {3D Coordinates};
\end{scope}

\end{tikzpicture}
\caption{\textbf{Matrix Representation of Synthesis Steps.} Each synthesis step concatenates reactant and product matrices along the row dimension, forming a $100\times64$ input. The columns encode three atomic attributes, five independent pharmacophore bits, a 50-column bond matrix, and six coarse/fine coordinate tokens. Only the final product in a route contains coordinates; reactants and intermediate products use padding in the coordinate columns. Reaction-template identifiers are not model inputs.}
\label{fig:matrix_rep}
\end{figure*}

For synthesis representation, a reaction step concatenates the reactant matrix
($50\times64$) and product matrix ($50\times64$) along the row dimension.
A route with $N$ reactions therefore has shape $(100N)\times64$. Steps are
ordered from the final reaction toward its precursors. Only the final product,
which occupies the product side of the first step, contains coordinate tokens;
all reactant and intermediate-product coordinate columns contain padding.

An intermediate is represented twice: as a fragment on the reactant side of
its parent reaction and as the product of its own reaction. This redundancy
allows the transformer to learn cross-step consistency. The model input and
output contain no reaction-template or tree-parent identifiers. Reaction
templates are assigned by replaying the generated reactants, and routes that
cannot be replayed or whose intermediates are not linked are rejected. The
result is a template-consistent synthesis sketch rather than a prediction of a
specific experimental reaction label.

\subsection{Discrete Diffusion Model}
\label{subsec:model}

We model the joint distribution of all categorical entries in a synthesis
route with an absorbing-MASK denoising objective related to discrete diffusion
models~\cite{austin2021structured}. Padding is an ordinary clean-data category;
MASK is a separate corruption token and is never a clean target. The model
generates reactants, products, topology, pharmacophore bits, and final-product
coordinates jointly, but does not generate reaction-template identifiers.

\subsubsection{Diffusion Process}
For each training route, we first sample a MASK probability $r$ from
\begin{equation}
    r \sim
    \begin{cases}
        \mathcal{U}(0,1), & \text{with probability }0.5,\\
        \mathcal{U}(0.95,1), & \text{with probability }0.4,\\
        1, & \text{with probability }0.1.
    \end{cases}
\end{equation}
For every matrix entry $j$, an independent mask indicator
$m_j\sim\operatorname{Bernoulli}(r)$ produces
\begin{equation}
    \widetilde{x}_j =
    \begin{cases}
        \texttt{MASK}, & m_j=1,\\
        x_j, & m_j=0.
    \end{cases}
\end{equation}
The denoiser predicts the clean category at each masked entry. Its training
objective is the mean cross-entropy over masked entries,
\begin{equation}
    \mathcal{L}(\theta) =
    -\frac{1}{\max(1,\sum_j m_j)}
    \sum_j m_j
    \log p_\theta(x_j\mid\widetilde{\mathbf{X}},\mathbf{c}),
\end{equation}
where $\mathbf{c}$ denotes the shape and pharmacophore conditions. The model
receives neither $r$ nor a discrete diffusion timestep; the MASK
pattern itself specifies which information is unavailable. The high-MASK
mixture emphasizes examples with little or no visible route-token context,
while the shape and pharmacophore conditions remain available.

Generation starts from an all-MASK route and performs a configurable number
$T$ of model evaluations. At iteration $k$, the model fills currently masked
entries and then remasks every logical position independently with
\begin{equation}
    r_{k+1}=1-\frac{k+1}{T}.
\end{equation}
Thus previously visible entries can be masked and revised later. We use
per-entry argmax proposals. Bond logits and sampled bond tokens are
symmetrized, and the two entries representing an undirected bond are always
remasked together. The final iteration has $r_T=0$ and returns a fully visible
route.

\subsubsection{Model Architecture}
\begin{figure*}[ht]
\centering
\begin{tikzpicture}[
    font=\sffamily\small,
    scale=0.85, transform shape,
    box/.style={
        draw=#1,
        fill=#1!8,
        thick,
        rounded corners=3pt,
        align=center,
        minimum height=0.9cm
    },
    smallbox/.style={
        draw=#1,
        fill=#1!8,
        thick,
        rounded corners=2pt,
        align=center,
        font=\scriptsize,
        minimum height=0.6cm
    },
    arrow/.style={
        ->,
        >=stealth,
        thick,
        #1
    }
]

\definecolor{inputblue}{HTML}{2E5AA8}
\definecolor{condgreen}{HTML}{2E8B57}
\definecolor{transpurple}{HTML}{7B68EE}
\definecolor{outputorange}{HTML}{D2691E}

\node[smallbox=inputblue, minimum width=1.6cm] (s1) at (0, 0) {Step 1};
\node[smallbox=inputblue, minimum width=1.6cm, right=0.3cm of s1] (s2) {Step 2};
\node[right=0.2cm of s2, font=\small, text=gray] (dots) {$\cdots$};
\node[smallbox=inputblue, minimum width=1.6cm, right=0.2cm of dots] (sn) {Step N};

\node[below=0.15cm of s2, font=\scriptsize, text=gray] {Synthesis Pathway ($L \times 64$)};

\node[smallbox=condgreen, minimum width=1.8cm, right=2.0cm of sn] (shape) {Shape};
\node[smallbox=condgreen, minimum width=1.8cm, right=0.5cm of shape] (pharm) {Pharmacophore};

\node[below=0.15cm of pharm, font=\scriptsize, text=gray, xshift=-0.9cm] {Conditions};

\node[box=transpurple, minimum width=7cm, above=1.2cm of s2, xshift=1cm] (encode) {
    \textbf{Feature Embeddings and Row Projection}\\[-2pt]
    \scriptsize $64 \times 16 \rightarrow H=1024$ + step and positional embeddings
};

\draw[arrow=inputblue!70] (s1.north) -- ++(0, 0.4) -| ($(encode.south west)!0.2!(encode.south east)$);
\draw[arrow=inputblue!70] (s2.north) -- ++(0, 0.5) -| ($(encode.south west)!0.4!(encode.south east)$);
\draw[arrow=inputblue!70] (sn.north) -- ++(0, 0.4) -| ($(encode.south west)!0.7!(encode.south east)$);

\node[box=transpurple, minimum width=14cm, minimum height=1.4cm, above=1.2cm of encode] (transformer) {
    \textbf{Transformer Encoder}\\[-2pt]
    \scriptsize 16 pre-normalized layers, 16 attention heads
};

\draw[arrow=transpurple!70] (encode) -- (transformer);

\node[smallbox=transpurple, minimum width=1.8cm, above=0.4cm of shape] (pool) {Mean + Max\\Pooling};
\draw[arrow=condgreen!70] (shape) -- (pool);
\draw[arrow=condgreen!70] (pool.north) -- (pool.north |- transformer.south);

\draw[arrow=condgreen!70] (pharm.north) -- ++(0, 1.8) -| ($(transformer.south west)!0.95!(transformer.south east)$);

\node[box=outputorange, minimum width=5cm, above=0.8cm of transformer] (project) {
    \textbf{Output Projection}\\[-2pt]
    \scriptsize $H \rightarrow 64 \times 102$
};

\node[above=0.5cm of project, align=center, font=\small] (output) {
    \textbf{Predicted $\mathbf{X}_0$}\\
    \scriptsize Denoised Tokens
};

\draw[arrow=outputorange!70] (transformer) -- (project);
\draw[arrow=outputorange!70] (project) -- (output);

\end{tikzpicture}
\caption{\textbf{Model Architecture.} The denoising network embeds the 64 categorical entries in each corrupted route row and projects their concatenation to $H=1024$. Step and positional embeddings provide route position. The shape point cloud is mean/max pooled to one token, whereas each pharmacophore point remains a separate token. A 16-layer, 16-head transformer jointly processes route and condition tokens and predicts the clean category at each route entry. The network has no explicit timestep or MASK-ratio input.}
\label{fig:network_architecture}
\end{figure*}

Our denoising model is a transformer encoder~\cite{vaswani2017attention} that
operates over route rows. It processes the corrupted route and conditioning
tokens as follows:

\begin{enumerate}
    \item \textbf{Feature Encoding}: Each of the 64 categorical entries uses a shared vocabulary of 102 tokens and is mapped to a 16-dimensional embedding. The 64 embeddings in a row are concatenated and projected to hidden dimension $H=1024$. Learned embeddings identify the reaction-step index and row position.
    \item \textbf{Condition Encoding}: The 200 shape points are encoded by a point-wise MLP, aggregated with both mean and max pooling, and projected to one transformer token. Each pharmacophore point is independently projected from its 8-dimensional feature vector and remains a separate token. Learned type embeddings distinguish the two condition types.
    \item \textbf{Row Processing}: Route and condition tokens are processed jointly by 16 pre-normalized transformer layers with 16 attention heads, GELU activations, a feed-forward dimension of $4H$, and zero dropout.
    \item \textbf{Output Projection}: Each route-row output is projected to $64\times102$ logits. Feature-specific vocabulary masks remove categories that are invalid for a given column before loss computation or sampling.
\end{enumerate}

\subsection{Conditional Generation}
\label{subsec:conditional}

We condition generation on ligand shape and atom-level pharmacophores. Both
signals are derived from the final product during training and from the
selected seed ligand during evolution; the network never receives protein
atoms, so training requires no protein--ligand complex structures. All stored
coordinates are centered on the final-product heavy-atom centroid. During
training, one random three-dimensional rotation is applied consistently to the
route coordinates, shape points, and pharmacophore coordinates.

\subsubsection{Shape Conditioning}
The ligand shape is represented by 200 unordered points. During training, each
point independently selects a final-product heavy atom with replacement and
adds a Cartesian offset sampled uniformly from $[-1,1]$~\AA{} along each axis.
No protein structure or distance filter is used to construct training
conditions.

During evolution, the shape condition is derived from the selected seed and
filtered against the receptor to avoid steric clashes with the binding pocket.
The resulting 200-point cloud is processed by the same shape encoder, which
combines mean and max pooling into one global condition token.

\subsubsection{Pharmacophore Conditioning}
Each pharmacophore point is an 8-dimensional vector containing five RDKit
features~\cite{landrum2013rdkit}---hydrogen-bond acceptor, aromatic,
hydrogen-bond donor, negatively ionizable, and positively ionizable---followed
by continuous Cartesian coordinates. The continuous condition coordinates
avoid requiring exact agreement with the 0.1~\AA{} generation grid.
Pharmacophore points receive no positional embeddings and are therefore
permutation-equivariant before transformer aggregation.

During training, only heavy atoms carrying at least one pharmacophore feature
are eligible. The complete pharmacophore condition is dropped with probability
0.1; otherwise, each eligible point is dropped with probability 0.5 and each
positive feature bit in a retained point is independently dropped with
probability 0.5. Coordinates of retained points are preserved even when all
of their feature bits are dropped. This incomplete conditioning allows the
model to introduce interactions not specified by the input.

During evolution, initial key atoms may be specified for a reference ligand.
For evolved seeds, key atoms can be selected from ligand atoms participating
in interactions detected by ProLIF (Section~\ref{subsec:scoring}). Their
features and coordinates form the pharmacophore condition for the next
generation.

Both conditions provide complementary guidance: shape biases generation toward
the seed's pocket-compatible envelope, while pharmacophore points bias the
model toward retaining selected interaction features without imposing a hard
structural constraint.

\begin{figure}[htbp]
    \centering
    \begin{minipage}{0.32\textwidth}
        \centering
        \includegraphics[width=\textwidth]{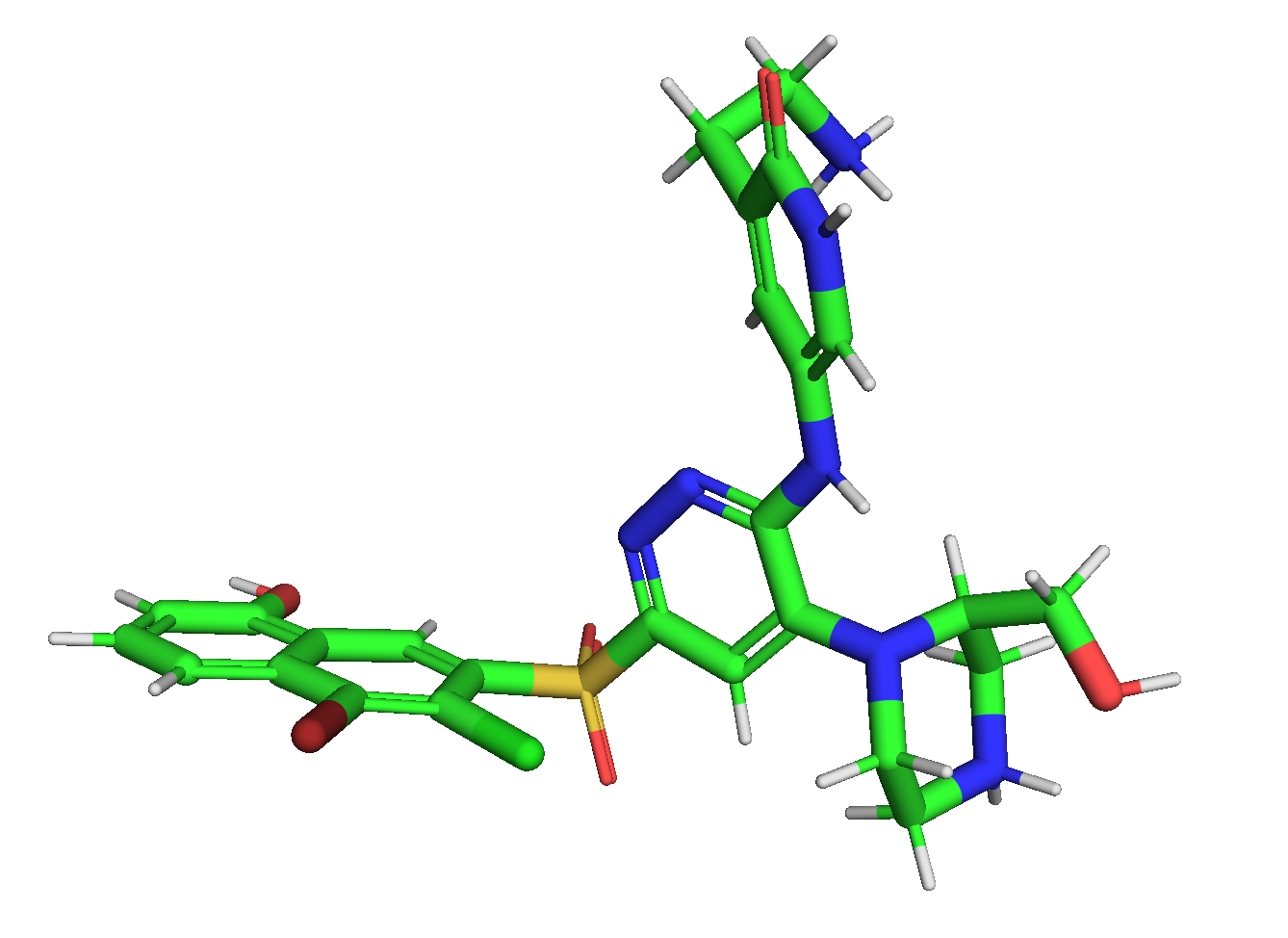}
        \centerline{\small (a) Ligand}
    \end{minipage}
    \hfill
    \begin{minipage}{0.32\textwidth}
        \centering
        \includegraphics[width=\textwidth]{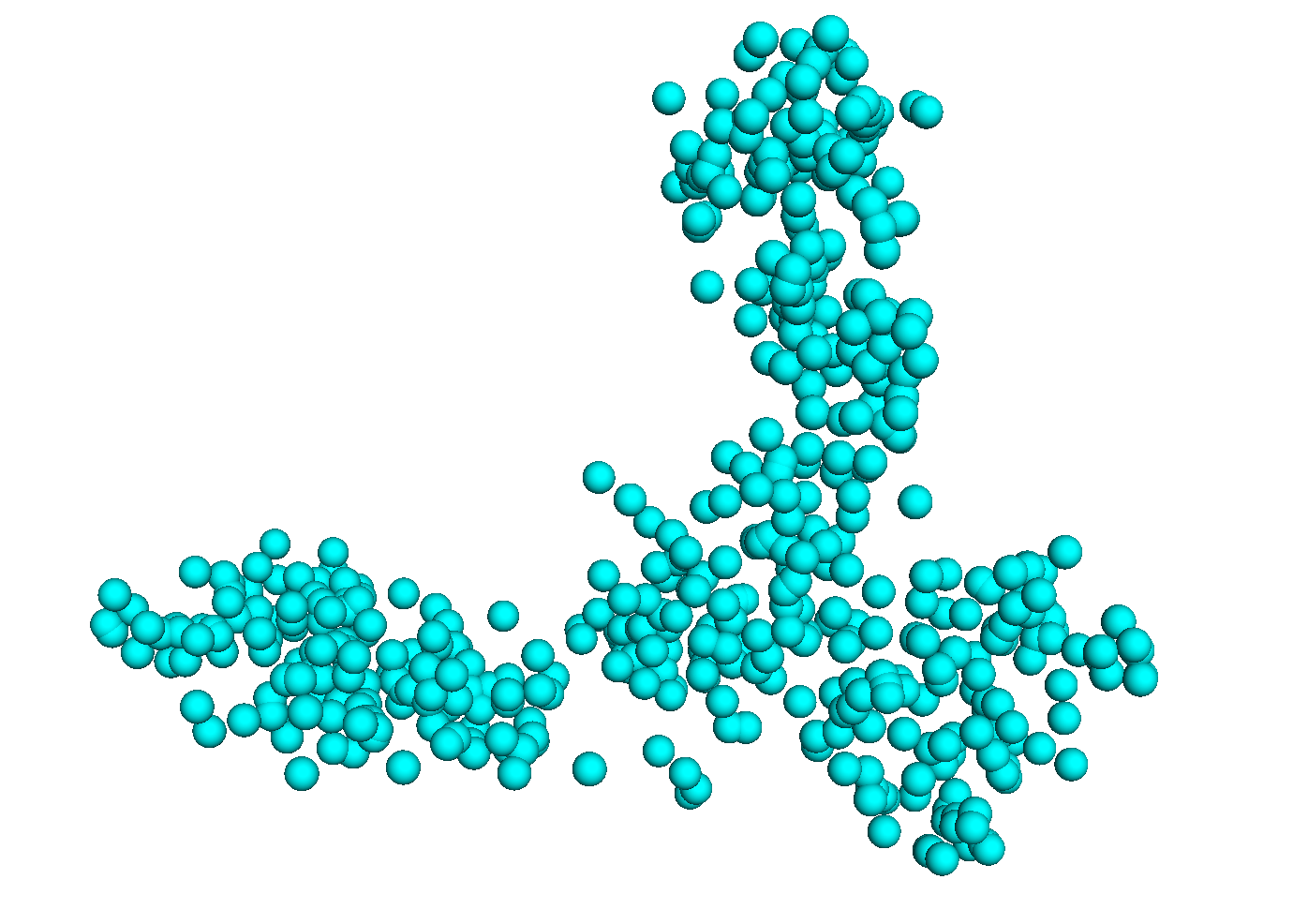}
        \centerline{\small (b) Shape Condition}
    \end{minipage}
    \hfill
    \begin{minipage}{0.32\textwidth}
        \centering
        \includegraphics[width=\textwidth]{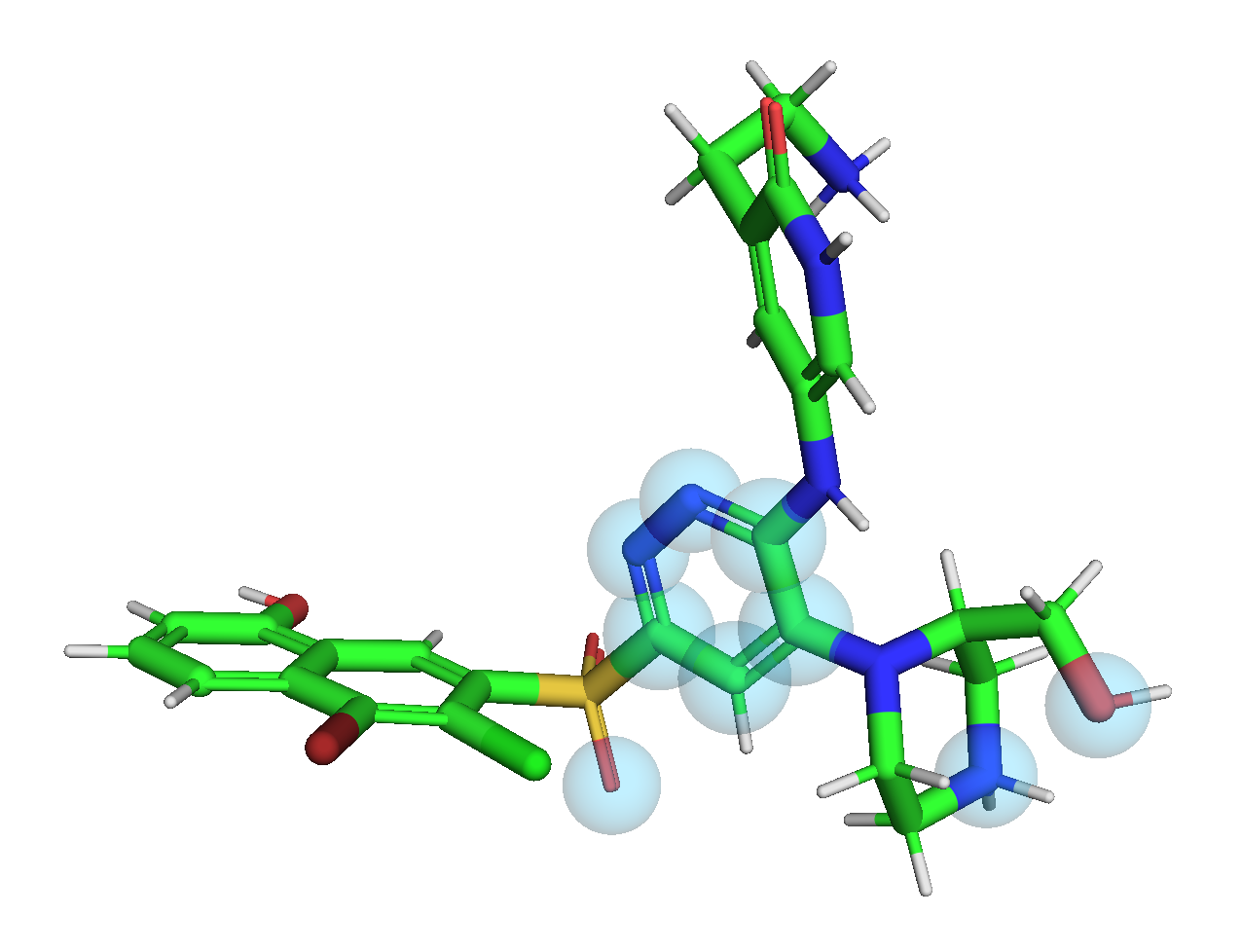}
        \centerline{\small (c) Pharmacophore}
    \end{minipage}
    \caption{\textbf{Conditioning Inputs.} (a) The ligand structure used for condition extraction. (b) The extracted shape condition represented as a point cloud. During evolution, points clashing with receptor atoms are removed. (c) The pharmacophore condition. Selected interaction-participating atoms (highlighted by translucent spheres) bias subsequent generation toward retaining those features.}
    \label{fig:conditioning_examples}
\end{figure}

\subsection{Molecular Evolution}
\label{subsec:evolutionary}

\begin{figure*}[ht]
\centering
\begin{tikzpicture}[
    font=\sffamily\small,
    scale=0.95, transform shape,
    block/.style={
        draw=#1,
        fill=#1!8,
        rounded corners=3pt,
        thick,
        align=center,
        minimum width=1.9cm,
        minimum height=1.0cm,
        font=\small
    },
    arrow/.style={
        ->,
        >=stealth,
        thick,
        #1
    }
]

\definecolor{inputblue}{HTML}{2E5AA8}
\definecolor{condgreen}{HTML}{2E8B57}
\definecolor{transpurple}{HTML}{7B68EE}
\definecolor{outputorange}{HTML}{D2691E}

\node[block=inputblue] (init) at (0,0) {Initialization};
\node[block=inputblue, right=0.6cm of init] (seed) {Seed Pool};
\node[block=transpurple, right=1.2cm of seed] (cond) {Condition\\Extraction};
\node[block=transpurple, right=0.6cm of cond] (gen) {Generation\\\& Filtering};
\node[block=transpurple, right=0.6cm of gen] (surr) {Surrogate\\\& ProLIF};
\node[block=outputorange, right=0.8cm of surr] (abfep) {ABFEP\\Score};

\draw[arrow=inputblue!70] (init) -- (seed);
\draw[arrow=transpurple!70] (seed) -- node[midway, above, font=\scriptsize, yshift=2pt] {Sample} (cond);
\draw[arrow=transpurple!70] (cond) -- (gen);
\draw[arrow=transpurple!70] (gen) -- (surr);

\draw[arrow=outputorange!70, dashed] (seed.north) -- ++(0,0.9) -| (abfep.north);

\path (surr.south) ++(0,-0.7) coordinate (turn1);
\path ([xshift=0.2cm]seed.south) coordinate (target1);
\path (abfep.south) ++(0,-1.1) coordinate (turn2);
\path ([xshift=-0.2cm]seed.south) coordinate (target2);

\path (turn2) -- coordinate[midway] (center_ref) (turn2 -| target2);

\draw[arrow=transpurple!70] (surr.south) -- (turn1) -- (turn1 -| target1) -- (target1);
\node[font=\scriptsize, above] at (center_ref |- turn1) {Select Seed};

\draw[arrow=outputorange!70, dashed] (abfep.south) -- (turn2) -- (turn2 -| target2) -- (target2);
\node[font=\scriptsize, below] at (center_ref) {Update Priority};

\begin{scope}[shift={(0, -2.4)}]
    \fill[transpurple!20, draw=transpurple, rounded corners=1pt] (-1.3, -0.12) rectangle (-0.9, 0.12);
    \draw[thick, transpurple!70] (-0.8, 0) -- (-0.2, 0);
    \node[font=\scriptsize, anchor=west] at (-0.1, 0) {Rapid Exploration (Inner Loop)};
    
    \fill[outputorange!20, draw=outputorange, rounded corners=1pt] (4.7, -0.12) rectangle (5.1, 0.12);
    \draw[thick, outputorange!70, dashed] (5.2, 0) -- (5.8, 0);
    \node[font=\scriptsize, anchor=west] at (5.9, 0) {Rigorous Verification (Outer Loop)};
\end{scope}

\end{tikzpicture}
\caption{\textbf{Molecular Evolution Process.} The framework utilizes a hierarchical evolutionary strategy. The inner loop (Rapid Exploration) iteratively generates, scores (Surrogate), and selects seeds to extract conditions for the next generation. Periodically, top candidates enter the outer loop for rigorous ABFEP verification, updating the seed pool with high-confidence binders.}
\label{fig:evolution_process}
\end{figure*}

We propose a generative molecular evolution strategy that progressively
refines molecules within the synthesizable chemical space. The framework
iterates generation, evaluation, and selection while retaining an explicit
synthesis route for every accepted candidate. Importantly, the loop requires
only a fitness signal and is not tied to a particular scoring implementation;
the energy functions used here are one instantiation for structure-based
design.

The evolutionary process unfolds in four phases:

\begin{enumerate}
    \item \textbf{Initialization}: For broad scaffold exploration, an initial population is generated using loose geometric conditioning derived from a reference ligand. For targets requiring specific interaction motifs, key pharmacophore constraints can additionally anchor the population around a desired binding mode.
    \item \textbf{Rapid Exploration}: A seed is sampled from the current population and converted into shape and pharmacophore conditions. The model generates molecules together with synthesis routes, after which project-specific filters and a computationally efficient fitness function rank the candidates. Promising molecules are retained as seeds, and interaction-participating atoms can provide pharmacophore conditions for subsequent generations.
    \item \textbf{Rigorous Verification}: Top-ranking candidates are periodically evaluated with ABFEP. Molecules with favorable binding free energies are selected as high-confidence seeds for the next evolution round, reducing error accumulation from the approximate rapid-exploration score.
    \item \textbf{Convergence}: Generation, evaluation, and selection continue until the chosen computational budget or project-specific stopping criterion is reached.
\end{enumerate}

Filters, property objectives, interaction constraints, and scoring models can
be adapted independently for each drug discovery project without changing the
synthesis-aware evolution loop.

\subsubsection{Rapid Exploration}
\label{subsec:scoring}
For the experiments in this work, we use chemical filters followed by a fast
surrogate energy score, while reserving ABFEP for outer-loop verification.

\textbf{Filtering}:
Generated molecules are screened for chemically unstable or undesirable
substructures using the SMARTS rules in
Appendix~\ref{sec:filtering_rules}. Target-specific physicochemical properties
and formal-charge preferences are incorporated as soft fitness penalties.
Protonation states are further checked before expensive
ABFEP evaluation.

\textbf{Molecular Relaxation}:
Although the model generates 3D coordinates, these coordinates primarily
specify the ligand conformation and provide geometric guidance for its binding
pose. We place each candidate in the receptor binding site, including any
retained cofactors, and perform a short energy minimization using
ff14SBonlysc~\cite{maier2015ff14sb} for the protein, ByteFF~\cite{zheng2024byteff}
for the ligand and cofactors, and OBC2 implicit
solvent~\cite{onufriev2004exploring}. This removes residual steric clashes and
provides a consistent structure for interaction detection and energy
evaluation.

\textbf{Interaction Constraints and Pharmacophore Mining}:
On the relaxed complex, ProLIF~\cite{bouysset2021prolif} detects
protein--ligand interactions such as hydrogen bonds and $\pi$--$\pi$ stacking.
Positive and negative interaction preferences can be incorporated into the
fitness with user-defined soft-penalty coefficients. For KRAS$^{\textrm{G12D}}$,
the absence of a salt bridge with ASP12 incurs a soft penalty. Ligand atoms
participating in these interactions can be reused as active pharmacophore
conditions in the next generation.

\textbf{Surrogate Energy Evaluation}:
In our experiments, the rapid-exploration score is based on minimized
potential energies in implicit solvent. We compute

\begin{equation}
    \Delta U = U_C - U_P - U_L
\end{equation}
where $U_C$ is the relaxed complex energy, $U_P$ is the receptor energy
evaluated at the same coordinates after removing the ligand, and $U_L$ is the
minimum ligand-only energy among 10 independently initialized minimizations.
More negative $\Delta U$ indicates a more favorable interaction. Optional
project-specific constraints are incorporated multiplicatively:
\begin{equation}
    S = \max(0, -\Delta U) \times \prod_i \lambda_i^{b_i}
\end{equation}
where $b_i\in\{0,1\}$ indicates violation of rule $i$ and
$\lambda_i\in(0,1]$ is its soft-penalty coefficient. The energy contribution
is set to zero when $\Delta U>0$.

While computationally efficient, this minimized implicit-solvent energy approach does not explicitly model entropic contributions and only approximates solvation and desolvation effects, which can be significant for hydrophobic pockets or water-mediated interactions. Therefore, this score serves primarily as a high-throughput filter to enrich the population with plausible binders before rigorous verification.

\subsubsection{Rigorous Verification}
For periodic verification, we apply Absolute Binding Free Energy Perturbation
(ABFEP)~\cite{zwanzig1954high} to top candidates from rapid exploration.
Unlike the surrogate score, ABFEP estimates binding free energy ($\Delta G$)
through dynamic sampling in explicit solvent. Candidates with the most
favorable $\Delta G$ are selected as seeds for the subsequent evolution round.
The same ABFEP protocol is used for the final comparison with reference
ligands.

Although the present experiments focus on binding affinity, the evolution
framework is extensible to other scalar or multi-objective fitness functions,
including ADMET, selectivity, and project-specific developability criteria.


\clearpage

\bibliographystyle{plainnat}
\bibliography{main}

\clearpage

\beginappendix

\section{Filtering Rules}
\label{sec:filtering_rules}

To ensure that the generated molecules are not only synthesizable but also chemically stable and suitable for drug discovery, we apply a set of rule-based filters. These rules are designed to remove molecules containing reactive electrophiles, unstable heteroatom bonds, specific structural alerts associated with toxicity, and known assay interference compounds (PAINS). The specific SMARTS patterns used for filtering are detailed below.

\subsection{Reactive Electrophiles}
Molecules containing highly reactive electrophilic groups are filtered out to prevent non-specific covalent modification of proteins (unless a covalent inhibitor is explicitly desired, which is not the case for the standard generation protocol).
\begin{itemize}
    \item \textbf{Aldehydes}: \verb|[CX3H1;!R;!$(C(=O)[N,O,S])]=!@[O;D1]| (Excludes formyl groups in stable contexts like amides).
    \item \textbf{Alpha-Halocarbonyls}: \verb|[C](=[O;D1])[C][Cl,Br,I]|.
    \item \textbf{Acyl Halides}: \verb|[C](=[O])[Cl,Br,I]|.
    \item \textbf{Sulfonyl Halides}: \verb|[S](=[O])(=[O])[F,Cl,Br,I]|.
    \item \textbf{Isocyanates \& Isothiocyanates}: \verb|[N]=[C]=[O,S]|.
    \item \textbf{Azides}: \verb|[$([N]=[N+]=[N-]),$([N-][N+]=[N])]|.
    \item \textbf{Cyanohydrins}: \verb|[C]([OH])C#N|
    \item \textbf{Epoxides \& Aziridines}: \verb|[C,N;r3]1@[O,N;r3]@[C,N,O;r3]1|.
    \item \textbf{Michael Acceptors}: \verb|[$([C]=[C]-[C,S,N](=[O,N,S])),$([C]=[C]-C#N)]|.
    \item \textbf{1,2-Dicarbonyl}: \verb|[#6,#8]C(=O)C(=O)[#6,#8]| (Reactive electrophiles, excludes stable oxalamides).
    \item \textbf{Alkyl Halides}: \verb|[CX4][Cl,Br,I]| (Alkylating agents, potential genotoxicity).
\end{itemize}

\subsection{Chemically labile groups}
We filter out specific chemical groups and linkages that are generally unstable or prone to decomposition under physiological conditions.
\begin{itemize}
    \item \textbf{N-N Single Bonds}: \verb|[#7&!a]-[#7&!a]| (Excluding aromatic N-N bonds like in pyrazoles).
    \item \textbf{N-O Single Bonds}: \verb|[#7&!a]-[#8&!a]| (Excluding aromatic N-O bonds like in Isoxazole).
    \item \textbf{O-O Single Bonds}: \verb|[#8]-[#8]|.
    \item \textbf{S-N / S-O Single Bonds}: \verb|[$([#16;!$(S(=O)=O)]-[#7]),$([#16;!$(S(=O))]-[#8])]| (Specific reactive sulfenamides and sulfenic esters are excluded).
    \item \textbf{Geminal Heteroatoms}: \verb|[N,O,S,P,Cl,Br,I]-!@[#6X4;!R]-!@[N,O,S,P,Cl,Br,I]|.
    \item \textbf{Heteroatom-Halogen Bonds}: \verb|[#7,#8,#15,#16]~[F,Cl,Br,I]|.
    \item \textbf{Carbamic Acids}: \verb|[#7]!@[C](=[O])!@[O;D1]|.
    \item \textbf{Anhydrides}: \verb|[C](=[O])[O][C](=[O])| (Unstable, prone to hydrolysis).
    \item \textbf{Pentafluorophenyl Esters}: \verb|C(=O)Oc1c(F)c(F)c(F)c(F)c1F| (Highly reactive activated esters).
    \item \textbf{Phosphorus-Heteroatom}: \verb|P~[N,S,Cl,C&D1]| (Unstable or toxic P-X bonds).
\end{itemize}

\subsection{Medchem Structural Alerts}
Certain functional groups are recognized as structural alerts for potential toxicity (e.g., mutagenicity) or metabolic liabilities by medicinal chemistry communities.
\begin{itemize}
    \item \textbf{Nitro Groups}: \verb|[N+](=O)[O-]|.
    \item \textbf{Thiocarbonyls}: \verb|[C]=[S]|.
    \item \textbf{Nitroso Groups}: \verb|[N;!$(N(=O)O)]=O|.
    \item \textbf{Azo Groups}: \verb|[#6]N=N[#6]| (Excluding aromatic N=N bonds).
    \item \textbf{Heavy Metals}: \verb|[Te,Se,Si,As,Sn,Hg,Fe,Pb,Pt]|.
    \item \textbf{Terminal Vinyls}: \verb|[CX3H2]=[CX3]|.
\end{itemize}

\subsection{PAINS \& Assay Interference}
We remove Pan-Assay Interference Compounds (PAINS) and other moieties known to cause false positives in screening assays.
\begin{itemize}
    \item \textbf{Catechols}: \verb|[O;H1,O-]-c:c-[O;H1,O-]|.
    \item \textbf{Thiophenols}: \verb|[S;H1,S-]-c|.
    \item \textbf{Quinones}: \verb|[$([O]=C1[C,N]=[C,N]C(=O)[C,N]=[C,N]1),$([O]=C1C(=O)[C,N]=[C,N][C,N]=[C,N]1)]| (Para- and Ortho-quinones).
    \item \textbf{S-S Single Bonds}: \verb|[#16]-[#16]| (Disulfides, often leading to assay artifacts).
\end{itemize}

\section{Evolutionary Experiment Settings}
\label{sec:evolutionary_settings_main}

\subsection{Hyperparameters}
\label{sec:evolution_settings}

To ensure robust sampling and avoid local optima, the evolutionary SBDD process was configured with the following hyperparameters for all targets:
\begin{itemize}
    \item \textbf{Seed pool size:} 10 molecules per evolutionary trajectory.
    \item \textbf{Generation batch:} 10 new molecules generated per step.
    \item \textbf{Parallel runs:} 8 independent evolutionary trajectories executed per target.
    \item \textbf{Diversity constraint:} Candidates sharing a Tanimoto similarity $> 0.5$ with any currently selected seed were explicitly excluded to prevent premature convergence to a single chemotype.
\end{itemize}

\subsection{Target-Specific Constraints}
\label{sec:target_constraints}

During the generative process, target-specific penalties were applied to steer the population toward viable drug candidates. The desired property ranges and structural constraints for each target are summarized in Table~\ref{tab:property_penalties}. These thresholds were empirically determined based on the physicochemical distribution of known active reference ligands for each target. All listed constraints are implemented as soft penalties, multiplying the fitness by $0.8$ for each violation.

\begin{table}[htbp]
    \centering
    \caption{Target-specific property and structural preferences applied during generation. Each entry denotes the desired range or interaction; each violation multiplies the fitness by a soft-penalty factor of $0.8$, without hard rejection. Entries marked ``--'' indicate no constraint.}
    \label{tab:property_penalties}
    \begin{tabular}{lcccccc}
        \hline
        \textbf{Property} & \textbf{KRAS$^{\textrm{G12D}}$} & \textbf{TYK2} & \textbf{BACE} & \textbf{c-MET} & \textbf{HSP90} & \textbf{p38} \\ \hline
        Molecular Weight (MW)                  & --             & $\le 500$      & $\le 450$    & $\le 520$    & $\le 500$    & $\le 550$ \\
        LogP                                   & $[-1, 5]$      & $[1.5, 5]$     & $[0, 5]$     & $[1.5, 4.8]$ & $[1, 5.5]$   & $[1, 5]$ \\
        Topological Polar Surface Area (TPSA)  & $\le 150$      & $\le 120$      & $\le 90$     & $\le 120$    & $\le 120$    & $\le 120$ \\
        H-bond Acceptors                       & $\le 10$       & $\le 10$       & $\le 10$     & $\le 10$     & $\le 10$     & $\le 10$ \\
        H-bond Donors                          & $\le 5$        & $\le 5$        & $\le 5$      & $\le 5$      & $\le 5$      & $\le 5$ \\
        Positive Charges                       & $\le 2$        & $0$            & $\le 1$      & $\le 1$      & $0$          & $0$ \\
        Negative Charges                       & $0$            & $0$            & $0$          & $0$          & $0$          & $0$ \\
        Salt bridge                            & ASP12          & --             & --           & --           & --           & -- \\
        \hline
    \end{tabular}
\end{table}

\section{Validation of ABFEP Calculations}
\label{sec:abfep_validation}

To rigorously evaluate the binding affinity of the generated molecules, we employed Absolute Binding Free Energy Perturbation (ABFEP). Before applying ABFEP to the generated candidates, we first validated its predictive accuracy on sets of known reference ligands for each of the six targets. The experimental binding affinities ($\Delta G_{\text{exp}}$) were compared against the ABFEP predictions to calculate several key performance metrics. The results are summarized in Table~\ref{tab:abfep_validation}.

\begin{table}[htbp]
    \centering
    \caption{Validation of ABFEP calculations on reference ligands across six targets. The metrics include Pearson correlation ($R$), Spearman's rank correlation ($\rho$), Kendall's $\tau$, Bias (mean signed error), centered RMSE (RMSE$_c$), and centered MAE (MAE$_c$). All error metrics (Bias, RMSE$_c$, MAE$_c$) are reported in kcal/mol.}
    \label{tab:abfep_validation}
    \begin{tabular}{lcccccc}
        \hline
        \textbf{Target} & \textbf{Pearson} & \textbf{Spearman} & \textbf{Kendall} & \textbf{Bias} & \textbf{RMSE$_c$} & \textbf{MAE$_c$} \\ \hline
        KRAS$^{\textrm{G12D}}$ & 0.736 & 0.842 & 0.689 & $-8.827$ & 4.563 & 2.883 \\
        TYK2 & 0.912 & 0.932 & 0.783 & $0.411$ & 0.516 & 0.396 \\
        BACE & 0.566 & 0.537 & 0.374 & $-1.482$ & 1.334 & 1.032 \\
        c-MET & 0.794 & 0.782 & 0.637 & $-4.211$ & 1.482 & 1.118 \\
        HSP90 & 0.893 & 0.830 & 0.689 & $1.680$ & 0.874 & 0.689 \\
        p38 & 0.810 & 0.829 & 0.666 & $-3.318$ & 0.750 & 0.551 \\
        \hline
    \end{tabular}
\end{table}

\section{Detailed Synthetic Routes for Generated Candidates}
\label{sec:synthesis_details}

In this appendix, we present the explicit computational synthesis sketches for the top-ranked candidates across the targeted protein families. For each molecule, the retrosynthetic plan details the small reactants and the sequence of reaction templates applied.

The routes provided here represent the raw output from MEVO's generative process. While they guarantee that the topological assembly of the molecule is consistent with known reaction patterns, they are abstract strategic disconnections. As discussed in Section~\ref{sec:results}, practical execution on the bench may require additional considerations, such as orthogonal protection/deprotection strategies, activation steps, and condition optimization, which are not explicitly modeled in these diagrams.

\begin{figure}[htbp]
    \centering
    \includegraphics[width=\textwidth]{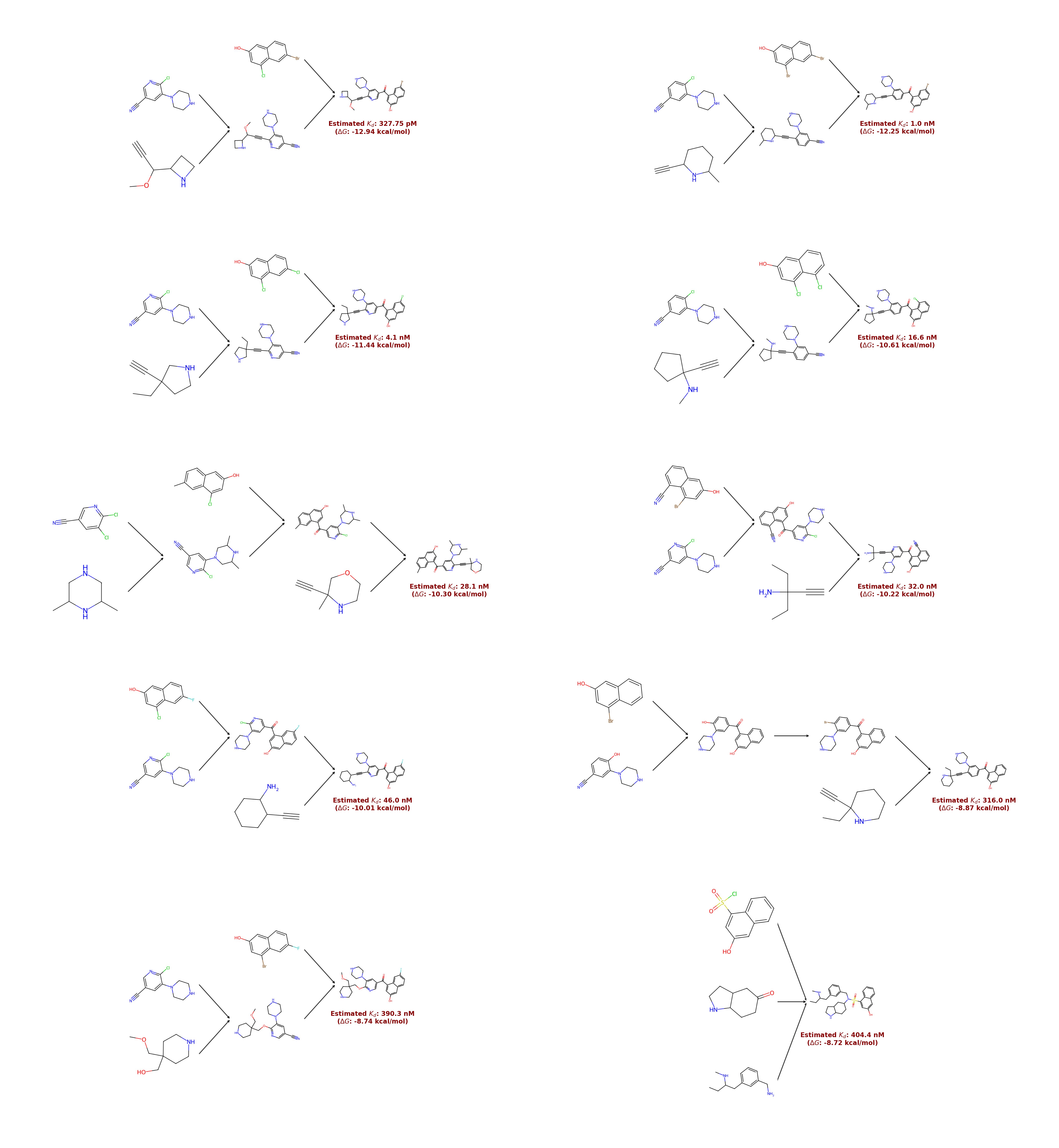}
    \caption{Synthetic routes for the top 10 MEVO-generated candidates targeting KRAS$^{\textrm{G12D}}$. The estimated $K_d$ and debiased $\Delta G_{bind}$ are shown for the final product.}
    \label{fig:synthesis_kras}
\end{figure}

\begin{figure}[htbp]
    \centering
    \includegraphics[width=\textwidth]{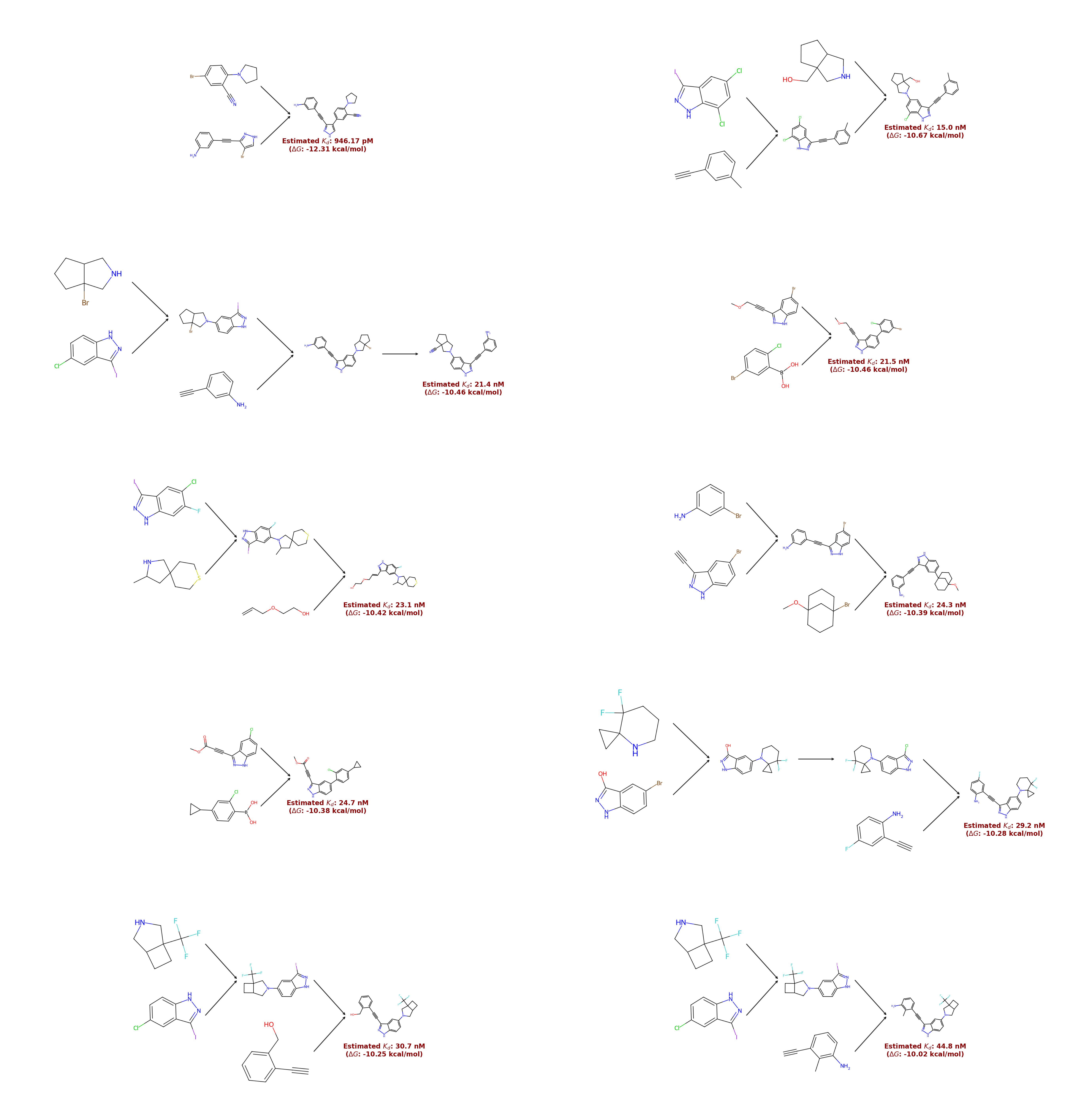}
    \caption{Synthetic routes for the top 10 MEVO-generated candidates targeting TYK2. The estimated $K_d$ and debiased $\Delta G_{bind}$ are shown for the final product.}
    \label{fig:synthesis_tyk2}
\end{figure}

\begin{figure}[htbp]
    \centering
    \includegraphics[width=\textwidth]{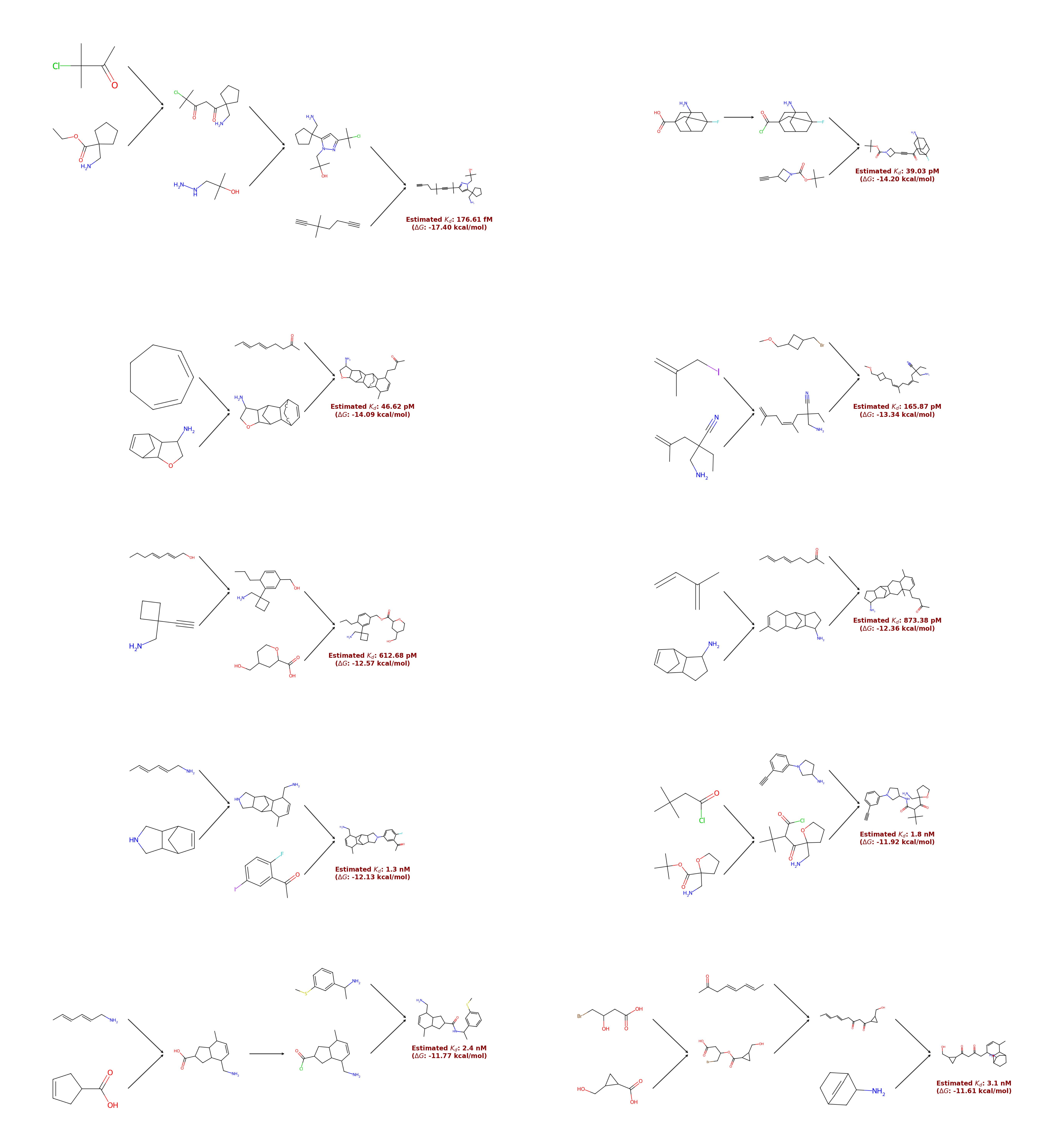}
    \caption{Synthetic routes for the top 10 MEVO-generated candidates targeting BACE. The estimated $K_d$ and debiased $\Delta G_{bind}$ are shown for the final product.}
    \label{fig:synthesis_bace}
\end{figure}

\begin{figure}[htbp]
    \centering
    \includegraphics[width=\textwidth]{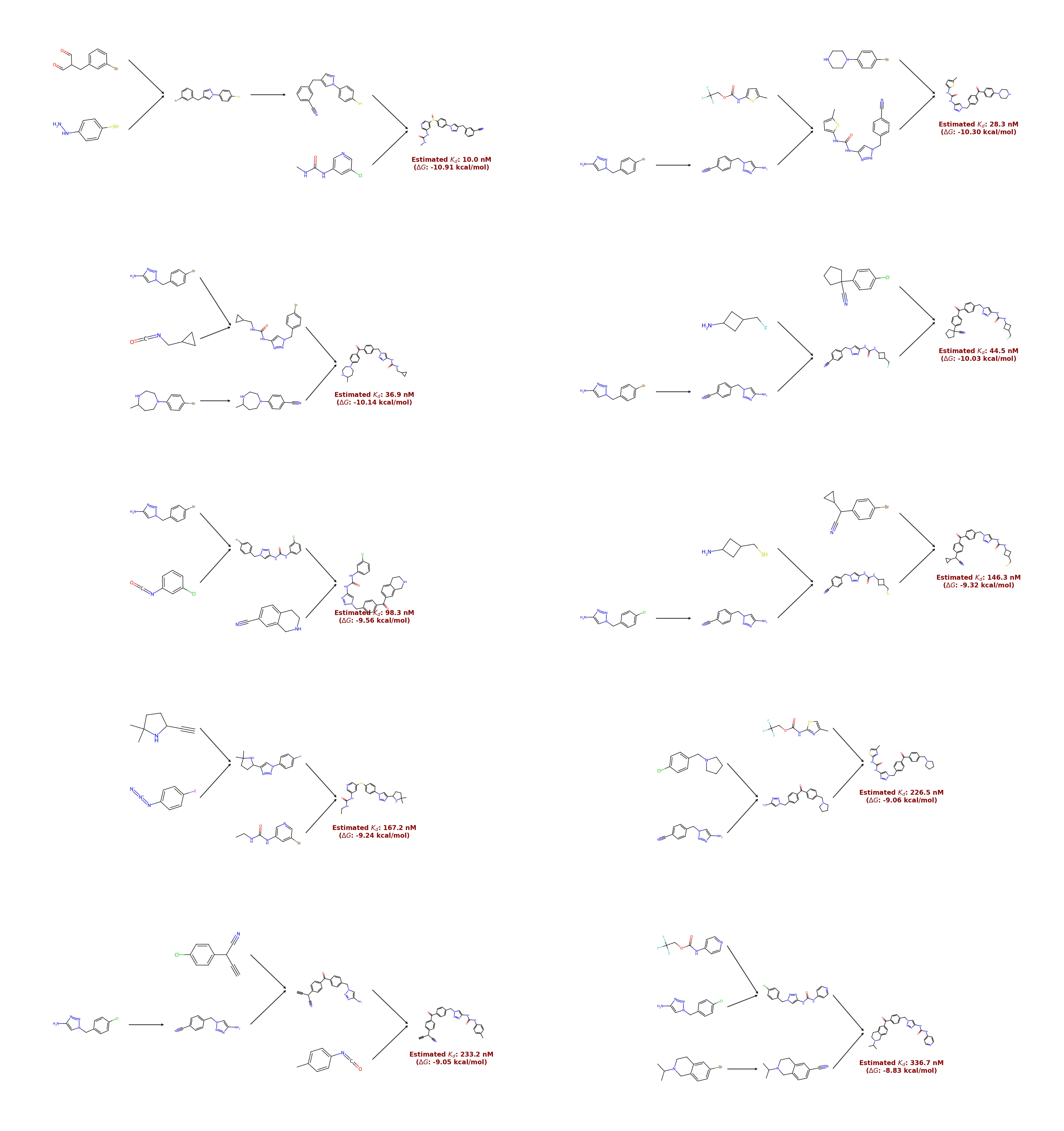}
    \caption{Synthetic routes for the top 10 MEVO-generated candidates targeting c-MET. The estimated $K_d$ and debiased $\Delta G_{bind}$ are shown for the final product.}
    \label{fig:synthesis_cmet}
\end{figure}

\begin{figure}[htbp]
    \centering
    \includegraphics[width=\textwidth]{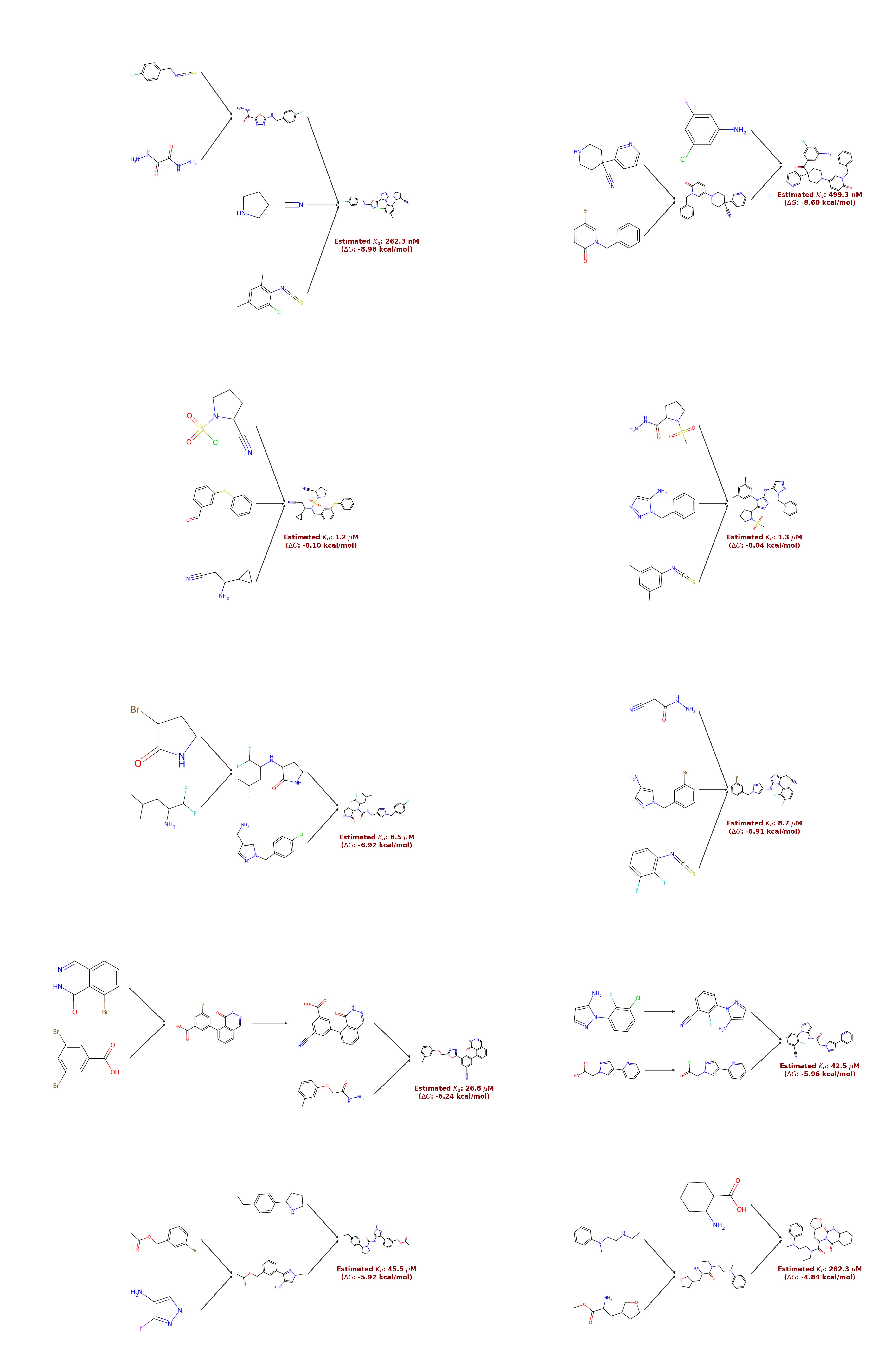}
    \caption{Synthetic routes for the top 10 MEVO-generated candidates targeting HSP90. The estimated $K_d$ and debiased $\Delta G_{bind}$ are shown for the final product.}
    \label{fig:synthesis_hsp90}
\end{figure}

\begin{figure}[htbp]
    \centering
    \includegraphics[width=\textwidth]{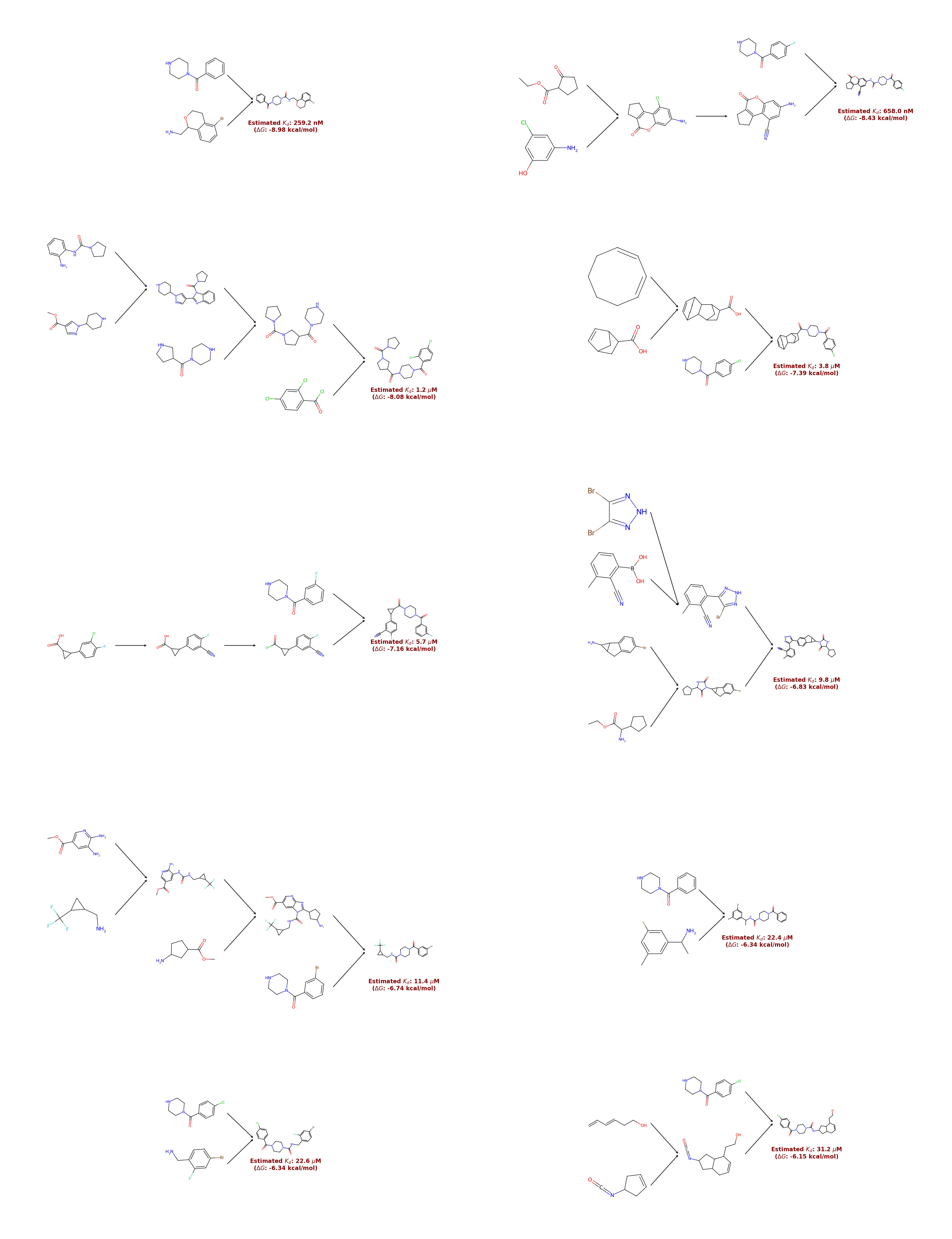}
    \caption{Synthetic routes for the top 10 MEVO-generated candidates targeting p38. The estimated $K_d$ and debiased $\Delta G_{bind}$ are shown for the final product.}
    \label{fig:synthesis_p38}
\end{figure}

\end{document}